\documentclass[lettersize,journal]{IEEEtran}
\usepackage{amsmath,amsfonts}
\usepackage{array}
\usepackage[caption=false,font=normalsize,labelfont=sf,textfont=sf]{subfig}
\usepackage{textcomp}
\usepackage{stfloats}
\usepackage{url}
\usepackage{verbatim}
\usepackage{graphicx}
\def\BibTeX{{\rm B\kern-.05em{\sc i\kern-.025em b}\kern-.08em
    T\kern-.1667em\lower.7ex\hbox{E}\kern-.125emX}}
\usepackage{balance}

\graphicspath{{pdf/}}
\usepackage{booktabs}
\usepackage{amssymb}
\usepackage{makecell}
\usepackage{multirow}
\usepackage{cite}
\begin{document}
\title{Reference-Guided Large-Scale Face Inpainting with Identity and Texture Control}

% \author{Wuyang~Luo\orcidlink{0000-0002-1447-3445}, Su~Yang, Weishan~Zhang
\author{Wuyang~Luo, Su~Yang, Weishan~Zhang
\thanks{This work was supported by State Grid Corporation of China (Grant No. 5500-202011091A-0-0-00). Corresponding author: Su~Yang.}
\thanks{Wuyang Luo and Su Yang are with the Shanghai Key Laboratory of Intelligent Information Processing, School of Computer Science, Fudan University, Shanghai 200433, China
(e-mail: wyluo18@fudan.edu.cn; suyang@fudan.edu.cn).} % <-this % stops a space
\thanks{Weishan Zhang is with School of Computer Science and Technology, China University of Petroleum, Qingdao 266580, China (e-mail: zhangws@upc.edu.cn).}
\thanks{Copyright © 2023 IEEE. Personal use of this material is permitted. However, permission to use this material for any other purposes must be obtained from the IEEE by sending an email to pubs-permissions@ieee.org.}}

% \markboth{Journal of \LaTeX\ Class Files,~Vol.~18, No.~9, September~2020}%
% {How to Use the IEEEtran \LaTeX \ Templates}

\markboth{IEEE Transactions on Circuits and Systems for Video Technology}
{How to Use the IEEEtran \LaTeX \ Templates}

\maketitle

% === ABSTRACT ====================================================================
% =================================================================================
\begin{abstract}
Face inpainting aims at plausibly predicting missing pixels of face images within a corrupted region. Most existing methods rely on generative models learning a face image distribution from a big dataset, which produces uncontrollable results, especially with large-scale missing regions. 
To introduce strong control for face inpainting, we propose a novel reference-guided face inpainting method that fills the large-scale missing region with identity and texture control guided by a reference face image.
However, generating high-quality results under imposing two control signals is challenging. 
To tackle such difficulty, we propose a dual control one-stage framework that decouples the reference image into two levels for flexible control: High-level identity information and low-level texture information, where the identity information figures out the shape of the face and the texture information depicts the component-aware texture. 
To synthesize high-quality results, we design two novel modules referred to as Half-AdaIN and Component-Wise Style Injector (CWSI) to inject the two kinds of control information into the inpainting processing. Our method produces realistic results with identity and texture control faithful to reference images. To the best of our knowledge, it is the first work to concurrently apply identity and component-level controls in face inpainting to promise more precise and controllable results. Code is available at \url{https://github.com/WuyangLuo/RefFaceInpainting}
\end{abstract}


% ====================================================================
% ====================================================================
% ====================================================================
%%%%%%%%% BODY TEXT
\section{Introduction}

\begin{figure}[t]
    \centering
    \includegraphics[width=8.5cm, trim=10 10 10 10,clip]{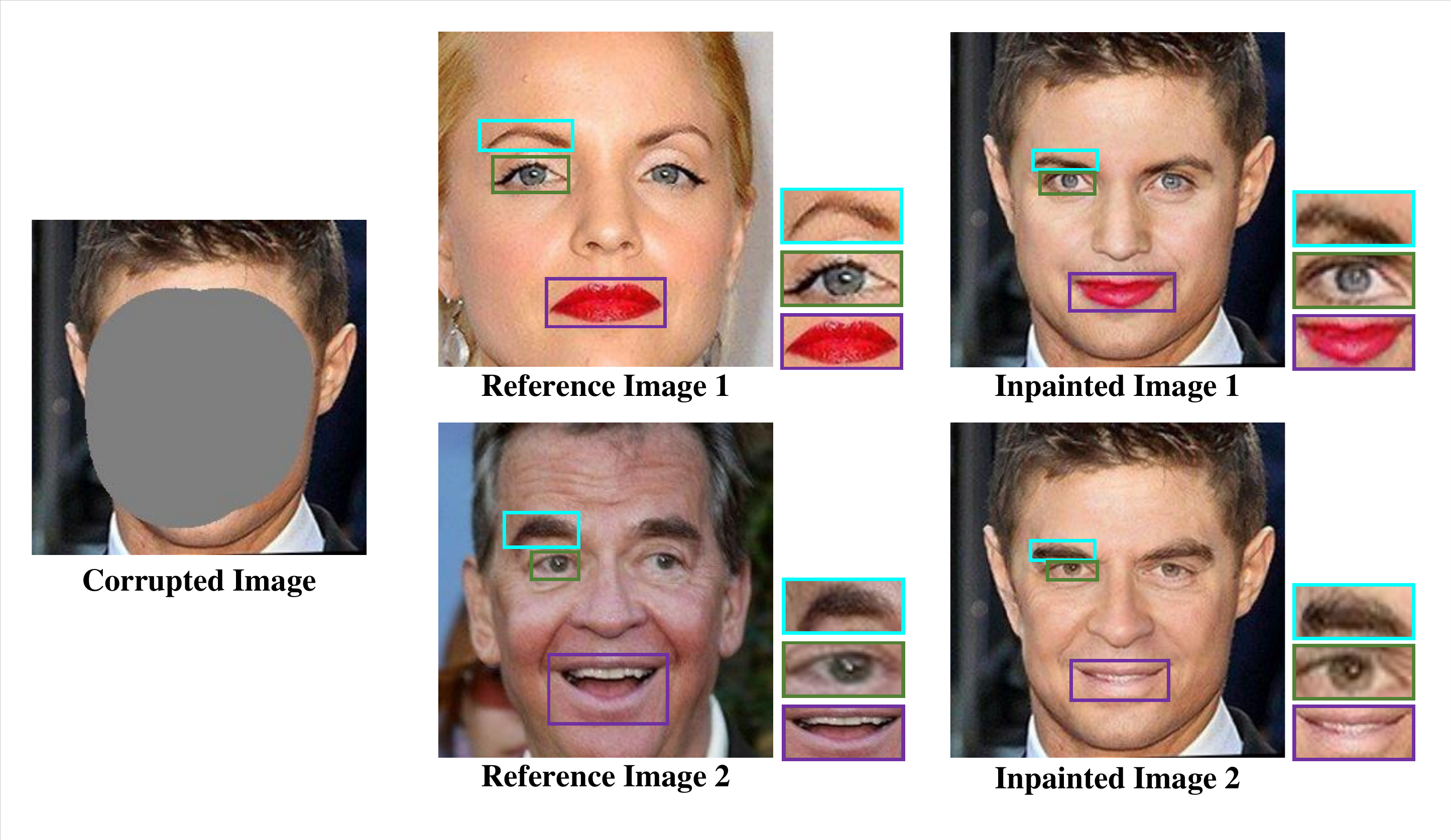}
    \caption{Our method introduces flexible control guided by reference images into face inpainting in terms of identity and texture.}
    \label{fig:intro}
\end{figure}

\IEEEPARstart{I}{mage} inpainting refers to filling up masked areas while keeping coherence with context. However, it is an ill-posed problem because a corrupted image can correspond to countless plausible completed results. Recent successful image inpainting methods rely on generative models learning a distribution from a big dataset and then predicting pixels of masked regions by known regions, which makes the generated content hold in a generic sense. This uncontrollability is unacceptable for face inpainting with large-scale missing regions. Because human eyes are extremely sensitive to face images, especially when the person in the photo is someone you know. In this case, inpainted results may be far from what the user desires. 
Recent works introduce a reference image to control the face inpainting process, called reference-guided face inpainting. It is a critical technique due to its wide applications, such as face frontalization \cite{duan2021simultaneous,tu2021joint}, restoration \cite{lahiri2020lightweight}, and generation \cite{xia2021local}. In this work, we concurrently introduce separate control of identity and texture into face inpainting, as shown in Fig. \ref{fig:intro}.

\begin{figure*}[t]
    \centering
    \includegraphics[width=17.5cm, trim=15 15 15 15,clip]{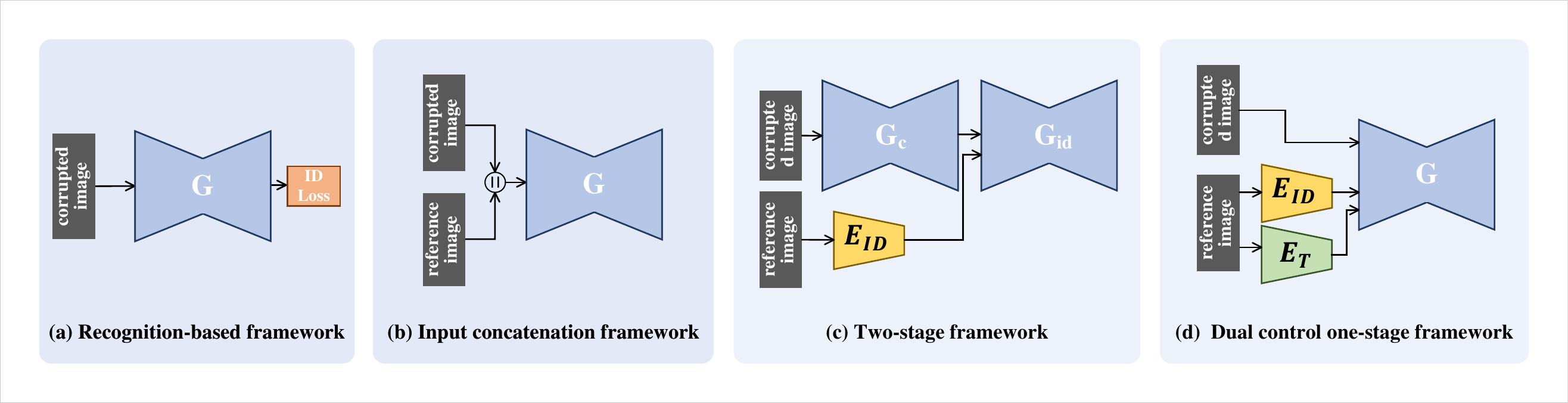}
    \caption{(a) Recognition-based framework: Only input the corrupted image and use identity loss on the output end to force the output's identity to be consistent with the identity of the ground truth. 
    (b) Input concatenation framework: Concatenate the reference image and the corrupted image together and send them into the generator.
    (c) Two-stage framework: Combine two methods of image inpainting and face swapping.
    (d) Dual control one-stage framework: We view our task as a whole and propose a one-stage framework that can control the identity and texture of results independently and concurrently. We utilize two encoders to extract identity and texture information from reference images, respectively.}
    \label{fig:schemes}
\end{figure*}

The existing works can be summarized into three frameworks, as shown in Fig. \ref{fig:schemes}. 
Recognition-based framework \cite{tcsvt} forces the identity of the inpainted image and the ground truth to be consistent using identity loss when training. This approach can only provide limited identity preservation. More importantly, it cannot freely control the identity of the output.
Input concatenation framework \cite{BMVC} concatenates the corrupted image and the reference image as input. This strategy always produces incongruous results due to discrepancies between corrupted and reference images in terms of color and pose.
Two-stage framework \cite{li2021swapinpaint} combines face inpainting and face swapping. It cascades two generators: The first generator is responsible for completing the face, and the second modifies the identity using identity features extracted from the reference image.
Input concatenation and two-stage frameworks can effectively control the high-level identity of the results, but they ignore low-level features of reference images, such as texture. Consequently, previous methods suffer from limited control and low-quality results. 

In this work, we propose a novel one-stage framework with dual control to concurrently exploit the high-level identity information and low-level texture information of reference images for controlling face inpainting and obtaining more realistic results. 
There are two key differences between our proposed approach and the existing methods. First, we decouple the reference image into high-level identity information and low-level texture style via two separate feature extractors to control the inpainting process. Identity information dominates the global structure and layout of large missing regions while texture information describes the diverse styles of face components, so this decoupled design enables our model to leverage control information from different perspectives and provides more flexibility. However, previous methods only consider identity information. 
Second, the two-stage framework splits the task into two separate steps: Face inpainting and face swapping, which leads to a larger number of parameters due to the fact that it requires two networks to achieve the goal. In contrast, our one-stage framework end-to-end optimizes our task within a neat model, achieving better performance with fewer parameters.

It is technically challenging to impose identity and texture control on an inpainting model independently and concurrently. For identity control, an identity vector can be injected into the generator via AdaIN \cite{Adain}, which is employed for face swapping \cite{chen2020simswap}. 
However, AdaIN is initially designed for the style transfer task, which is proven to wash away the original style of feature maps \cite{Adain,luo2022context}, such as colors and textures. For our image inpainting task, the generated regions must be style-consistent with the known regions. Applying AdaIN directly to the entire feature map may result in inconsistent styles, as shown in Fig. \ref{fig:ablation}. Therefore, we propose a light-overhead but effectively improved module referred to as Half-AdaIN, where we let half of the feature maps go through the AdaIN operation, and the remaining channels bypass AdaIN to avoid completely washing away the style of the input image.

For texture control, we design a Component-Wise Style Injector (CWSI) to locate, extract, and inject texture information of pre-defined face components. Precisely, CWSI first extracts the style codes of the reference image using a region-wise style encoder. 
Then, CWSI parses the feature maps to obtain the intermediate-level segmentation map for locating face component regions. 
Finally, CWSI injects style codes into corresponding component regions. To improve the texture control performance, we use a three-mode training scheme to approach better overall results.
Our contributions are summarized as follows:
\begin{itemize}
\item[$\bullet$] We introduce precise separate control of identity and texture into face inpainting, for which we propose a novel dual control one-stage framework for reference-guided face inpainting.
\item[$\bullet$] We design two new modules: Half-AdaIN and Component-Wise Style Injector (CWSI). They effectively inject identity and texture information to impose fine-grained controls on global profiles and local details simultaneously.
\item[$\bullet$] Extensive experiments are performed to show that our method outperforms the state-of-the-art methods and can provide flexible control subject to reference images in terms of identity and texture.
\end{itemize}

\begin{figure*}[t]
    \centering
    \includegraphics[width=17.8cm, trim=20 10 10 10,clip]{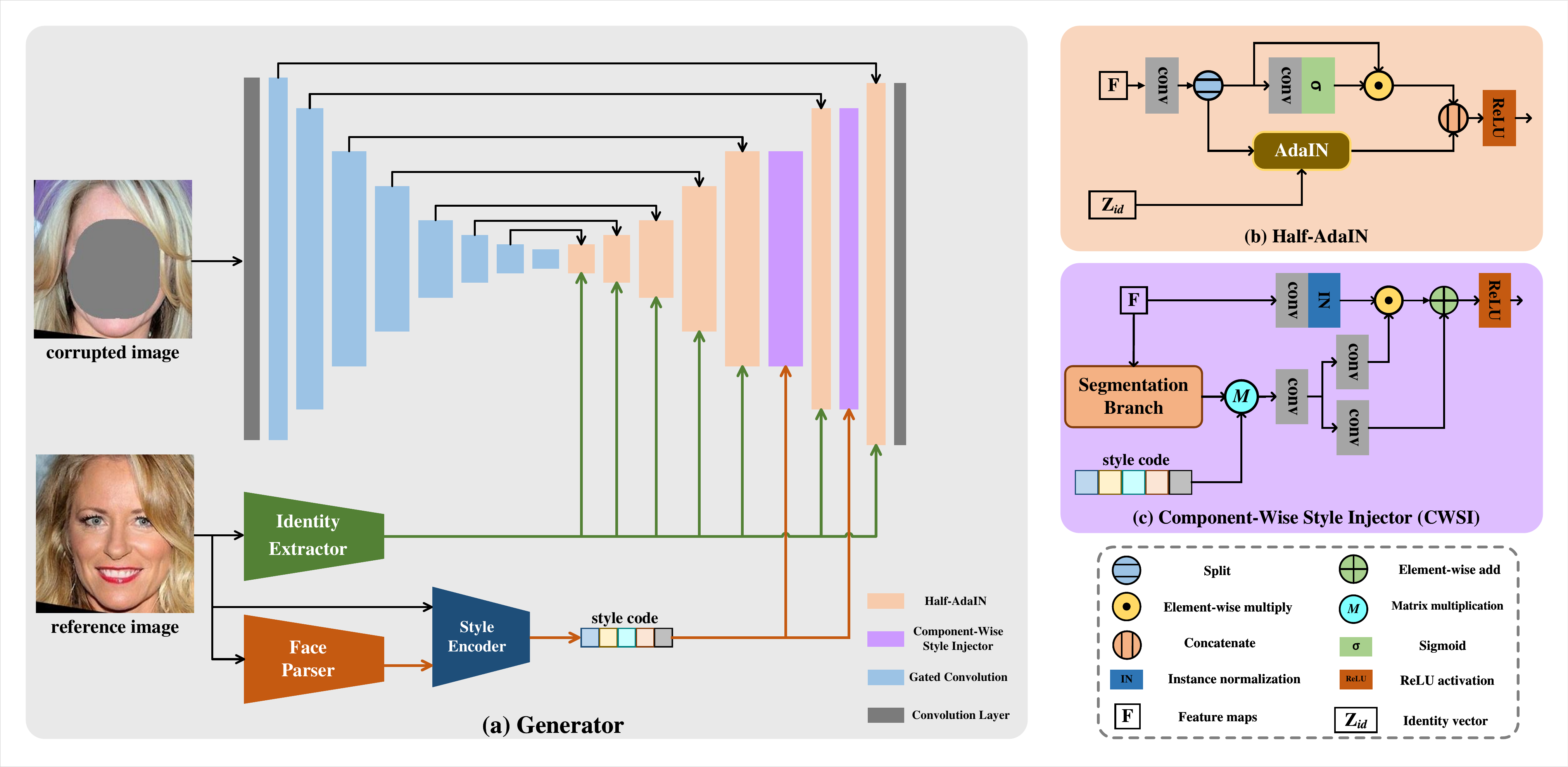}
    \caption{Overview of the proposed framework. (a) Generator; (b) Half-AdaIN; (c) Component-Wise Style Injector (CWSI).}
    \label{fig:framework}
\end{figure*}

\section{Related Works}
\subsection{Image Inpainting}
The existing inpainting methods can be divided into two categories: Traditional methods and deep learning based methods. Traditional methods use the remaining pixels to fill missing pixels, including diffusion-based methods \cite{ballester2001filling, bertalmio2000image, levin2003learning} and patch-based methods \cite{efros1999texture, kwatra2005texture, efros2001image}. These methods perform well on stationary textures. However, they often fail to synthesize satisfactory results for complex scenes because they do not understand high-level semantics. 
The methods based on deep learning can learn to understand the high-level semantics of images by training neural networks on large-scale datasets. Recent works have been devoted to improving image inpainting performance by introducing adversarial training \cite{GAN,pathak2016context, iizuka2017globally}, contextual attention mechanism \cite{CA, zeng2019learning,zeng2021cr,liu2022deep,wang2022dual}, new convolutions scheme \cite{GC,pconv}, additional information \cite{nazeri2019edgeconnect, ren2019structureflow,xu2020e2i,wang2020structure}, diversity  \cite{zheng2019pluralistic, liu2021pd}, memory \cite{feng2022generative}, and vision transformer \cite{wan2021high}.

As a prevalent branch of image inpainting, face inpainting has benefited a lot from image inpainting. 
Face inpainting methods often use rich prior knowledge or external conditions of human faces to impose constraints improving performance, such as semantic structure \cite{li2017generative}, attributes \cite{xiao2021face}, example images \cite{eye, BMVC}, identity \cite{li2020learning}, symmetry \cite{tcsvt}, and correlation among facial parts \cite{zhou2020learning}.
However, the existing methods only apply distribution learned from big datasets to direct inpainting, so the global structure as well as fine details are in general out of control, and not able to fit well into any specific desire.

\subsection{User-Guided Image Inpainting}
Some works introduced user interaction based clues to guide inpainting for control, such as segmentation maps \cite{luo2022context}, line \cite{sun2005image}, transformation \cite{pavic2006interactive}, image library  \cite{hays2007scene}, structures \cite{huang2013transformation}, color \cite{jo2019sc}, and text \cite{zhang2020text}. Moreover, introducing another image as a reference is an intuitive way to exert comprehensive control. However, few works \cite{Transfill,BMVC,eye,li2021swapinpaint} have been developed to make use of all the potential clues from the reference image  \cite{Transfill,BMVC,eye}. \cite{eye} only considers how to inpaint a specific component within a small local area and does not care about the global structure. \cite{BMVC} only inserts reference images at the beginning of the generator and cannot control finely the final outcome. \cite{Transfill} requires the reference image and the inpainted image to be consistent in content, lighting, and view, which is impractical in many scenarios. \cite{li2021swapinpaint} only considers high-level identity consistency but neglects local texture information. This paper proposes a new framework for reference-guided face inpainting tasks to make full use of the control information provided by the reference image in terms of identity and texture. 

\subsection{Conditional Face Generation}
Conditional face generation attempts to synthesize face images based on input conditions. It contains various subtasks, such as face image translation \cite{luo2022photo, pix2pix, HDpix2pix,SPADE,SEAN}, face swapping \cite{chen2020simswap,li2019faceshifter}, talking head synthesis \cite{jamaludin2019you,prajwal2020lip}, face frontalization \cite{duan2021simultaneous,duan2020look,tu2021joint}, and face stylization \cite{liu2021blendgan}. Our work is related to two previous methods \cite{chen2020simswap,SEAN}. SimSwap \cite{chen2020simswap} utilizes the AdaIN operation to build its generator, which transfers the identity information of the source face into the target face at feature level for arbitrary face swapping. To inject identity features while preserving the other important original features, we propose an improved version of AdaIN, referred to as Half-AdaIN. SEAN \cite{SEAN} synthesizes face images from segmentation maps, which can control the style of each semantic region individually. They utilize a style encoder network to extract the style of each region. 
We reuse their design to extract style codes from reference images and then further employ the proposed CWSI for component-aware style injection serving face inpainting.

The proposed method introduces controllability to face inpainting so it can achieve an effect that looks similar to face swapping. However, our face inpainting task fundamentally differs from face swapping: 
(1) Different input: The input of our task is a face image with an arbitrary hole. Therefore, the structure of the input face is incomplete, e.g., missing nose or mouth; the input of face swapping is a complete face image. 
(2) Different goals: Face inpainting aims to reconstruct the missing region to match the known region. For example, if the central region of the input face is invisible, the face inpainting method needs to generate components of the face, such as the nose, at appropriate locations. Face swapping aims to change the overall identity of the input to another identity. Face swapping can change all identity-related elements, such as the shape of the nose and the size of the eyes. But face swapping does not need to reconstruct the structure.
(3) Different outputs: Face inpainting fills the hole and generates a complete face. However, the content inside the hole is to some extent uncontrollable, e.g., if the upper part of the face falls into the missing region, the position and shape of the generated eyes may be different if using different face inpainting models. Face swapping does not need to reconstruct the structure of the face, so its output and input are aligned in terms of position and pose.

\section{Method}
We describe our approach in a top-down manner. We first introduce the architecture of our generator and then give details of two novel modules for identity and texture control. 

Our generator has two inputs: (1) The corrupted image with the corresponding mask whose value is 0 in the missing region and 1 in the known region;  (2) The reference image provides control information that users desire.
The generator adopts an encoder-decoder architecture with skip connections \cite{unet}, as illustrated in Fig. \ref{fig:framework}(a). Specifically, the encoder is composed of several successive gated convolutional layers \cite{GC} with stride 2. The decoder is composed of several Half-AdaIN with upsampling operations. In addition, two CWSI are inserted into the decoder at the resolution of $64 \times 64$ and $128 \times 128$. For the inpainting task, we generally believe that the encoder can transfer the available information from the known pixels to missing pixels by gradually increasing the receptive field, and the decoder is responsible for the reconstruction of details  \cite{yan2018shift}, so we only place the control modules in the decoder.

In order to force the training generator to produce more realistic outputs, we employ a global discriminator and three local discriminators against the generator. We use SN-PatchGAN \cite{GC} as our global discriminator. The local discriminators are focused on specific sub-regions, including two eyes and mouths, which helps the generator synthesize high-frequency textures in these regions. The discriminators are omitted in Fig. \ref{fig:framework}.

\begin{figure*}[t]
    \centering
    \includegraphics[width=17.0cm, trim=10 10 10 10,clip]{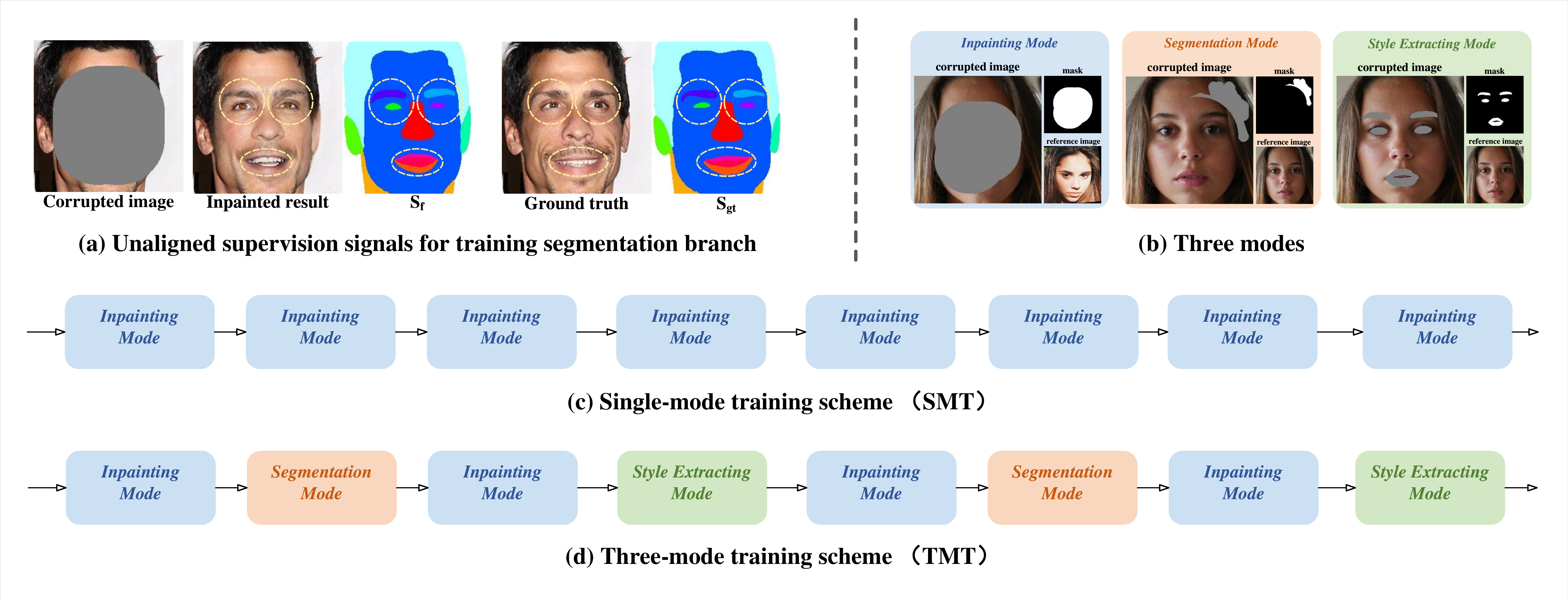}
    \caption{(a) Misalignment issue of supervision signals for segmentation branches. The yellow circles enclose the unaligned regions between inpainted results and ground truth. (b) Different masks and reference images are applied for different modes. (c) Single-mode training scheme. (d) Three-mode training scheme.}
    \label{fig:training_mode}
\end{figure*}

\subsection{Identity Control}
% \noindent
% \textbf{Identity feature extraction}

\emph{1) Identity Feature Extraction: }
We employ a well-trained face recognizer model \cite{deng2019arcface} as the identity feature extractor. We take the output of the last layer prior to the classifier and normalize this 512-dimensional embedding as the identity vector $Z_{id}$.
Compared to using reference images as guidance directly \cite{BMVC}, the high-level identity embedding extracted from the recognition model trained on large-scale face datasets can figure out more representative identity features. The identity embedding largely ignores pose, background, and lighting. Thus, it is unnecessary to require rigorous spatial alignment between the corrupted image and the reference image.

\begin{figure*}[t]
    \centering
    \includegraphics[width=18.0cm, trim=20 10 130 10,clip]{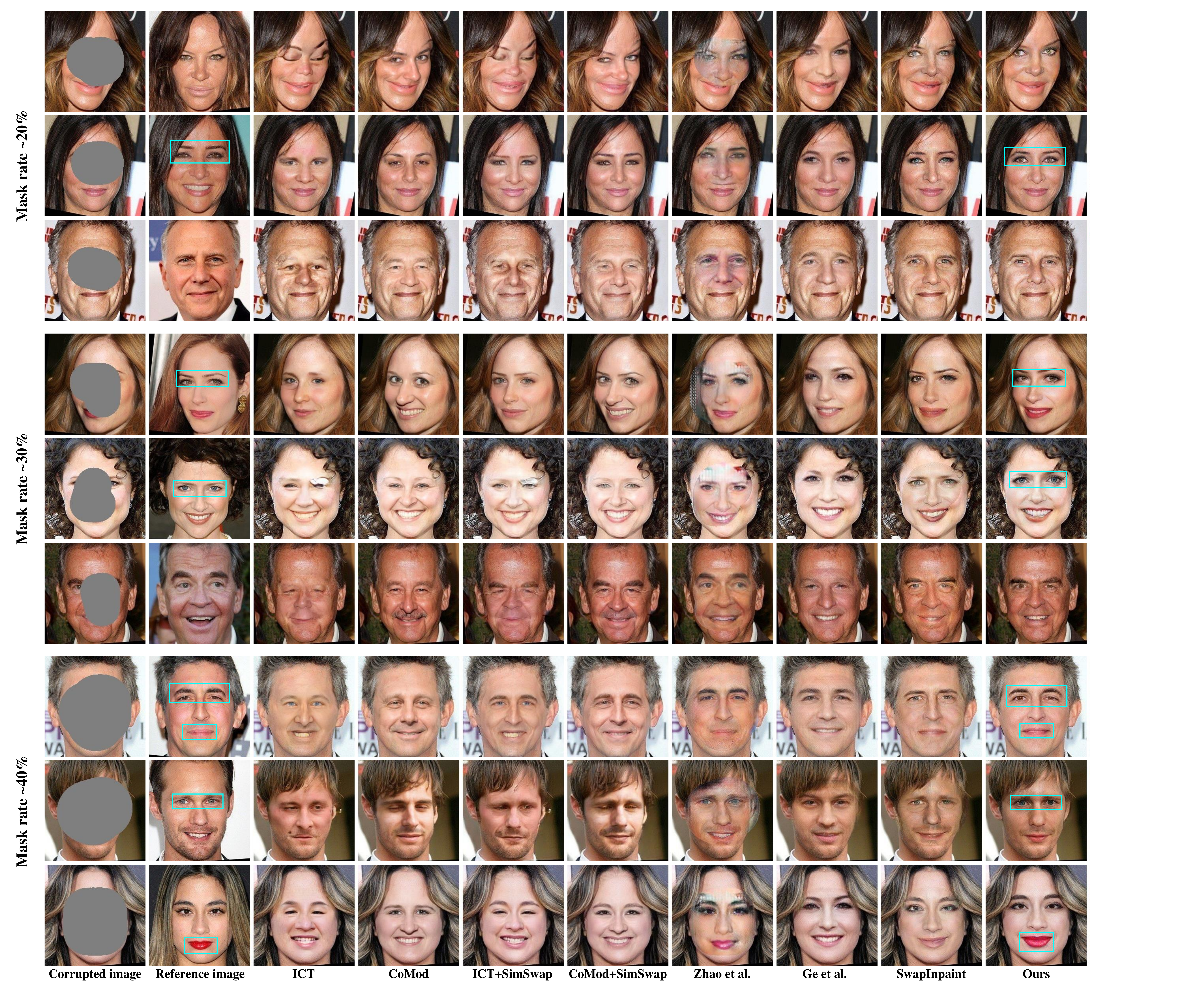}
    \caption{Visual comparison with baselines. The turquoise lines highlight that our method can synthesize texture-consistent results.}
    \label{fig:sota}
\end{figure*}

% \noindent
% \textbf{Half-AdaIN for Identity Injection}

\begin{figure}[t]
    \centering
    \includegraphics[width=8.5cm, trim=10 10 130 10,clip]{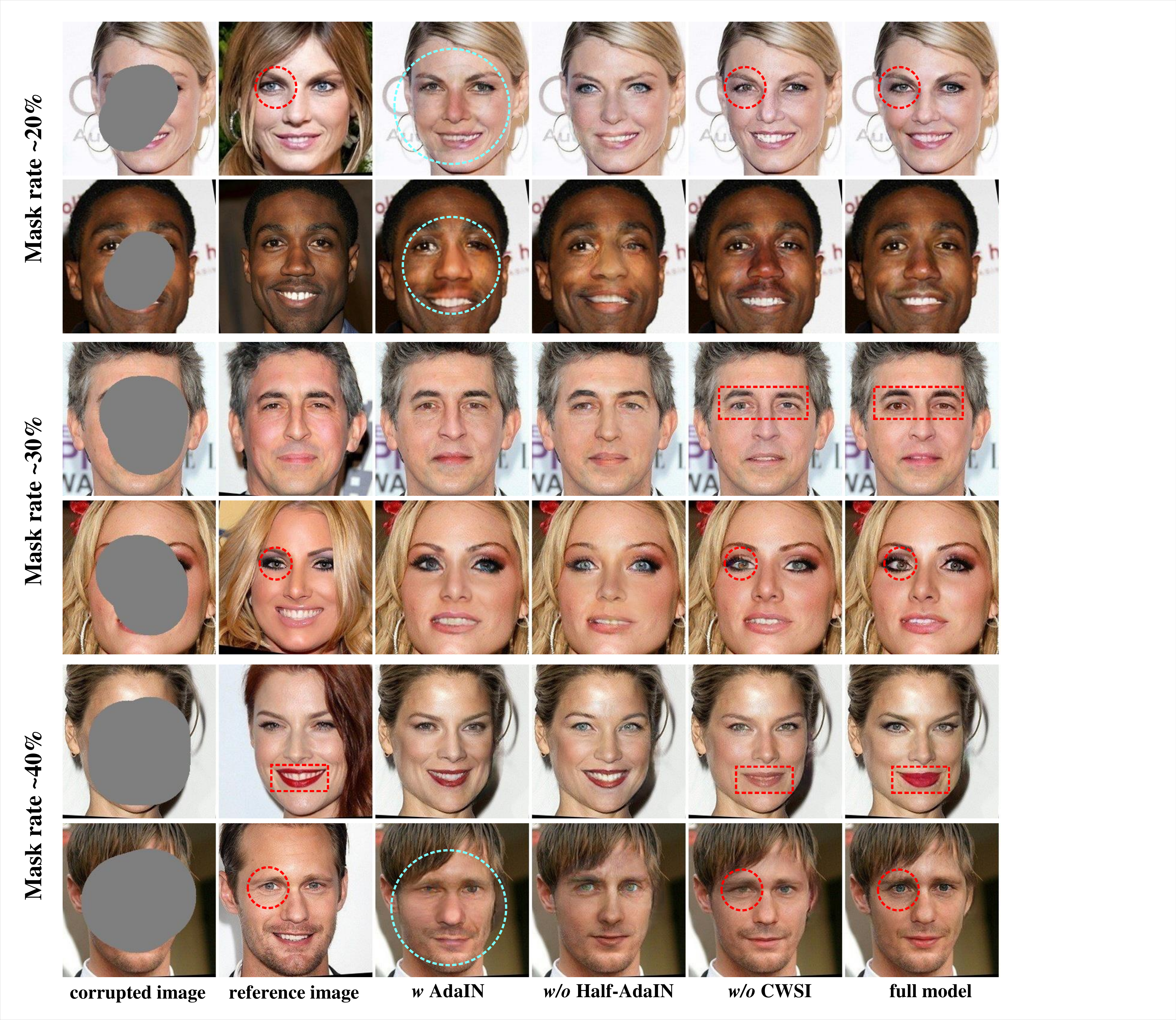}
    \caption{Visual results of ablation studies. The red dotted lines point out the texture inconsistency with the reference images. The turquoise dotted lines enclose the low-quality results.}
    \label{fig:ablation}
\end{figure}

\emph{2) Half-AdaIN For Identity Injection: }
The previous face swapping work \cite{chen2020simswap} borrows the AdaIN \cite{Adain} operation to inject the identity feature. However, if we naively reuse AdaIN for our task, it produces many low-quality results as shown in Fig. \ref{fig:ablation}. AdaIN washes away the original style of feature maps revealed by \cite{Adain}, so it is inappropriate for our task that requires injecting an external identity vector while keeping coherence between generated content and known pixels. To tackle this problem, we propose a simple but effectively improved module referred to as Half-AdaIN, as illustrated in Fig. \ref{fig:framework}(b). The key novelty is that we split the input feature maps for different cues, half for identity control and the other half for preserving the original contextual information.

Let $F \in \mathbb{R}^{C \times H \times W}$ and $Z_{id} \in \mathbb{R}^{C_{id}}$ denote input feature maps and the identity vector, respectively. Here, $C$ is the number of channels, and $H$ and $W$ represent spatial dimensions.
First, $F$ passes through a standard convolutional layer to obtain $\widehat{F} \in \mathbb{R}^{C \times H \times W}$. Then, we split $\widehat{F}$ into two equal slices along the channel dimension. Thus, we get two tensors $F_{1} \in \mathbb{R}^{C/2 \times H \times W}$([$0: \frac{C}{2}$) channels) and $F_{2} \in \mathbb{R}^{C/2 \times H \times W}$([$\frac{C}{2}: C$) channels).  We inject identity information by applying a standard AdaIN to $F_{1}$. Specifically, first, $F_{1}$ is convolved to $\widehat{F_{1}}$, and then we perform instance normalization on $\widehat{F_{1}}$:
\begin{equation}
\bar{F_{1}}=\frac{\widehat{F_{1}}-\mu}{\sigma}
\end{equation}
where ${\mu} \in \mathbb{R}^{C/2}$ and ${\sigma} \in \mathbb{R}^{C/2}$ are the channel-wise means and standard deviations of $\widehat{F_{1}}$. 
Then, we modulate feature maps with scaling parameters $\gamma$ and shifting parameters $\beta$ learned from identity embedding $Z_{id}$. The modulation process is formulated as:
\begin{equation}
\widetilde{F_{1}}= \gamma \odot \bar{F_{1}} \oplus \beta
\end{equation}
where ${\gamma} \in \mathbb{R}^{C}$ and ${\beta} \in \mathbb{R}^{C}$ are generated from $Z_{id}$ through two fully connected layers and expanded to match the spatial resolution of $\bar{F_{1}}$. $\odot$ and $\oplus$ are element-wise multiplication and addition,  respectively. 
For $F_{2}$, to adaptively distinguish valid and invalid pixels during the inpainting process, we apply a spatial attention mechanism following gated convolution \cite{GC}. Specifically, we generate a soft weight map $W$ using $F_{2}$ through a convolutional layer followed by a sigmoid operation to make the values of $W$ between 0 and 1. Finally, the output of the Half-AdaIN is formulated as: 
\begin{equation}
F_{out}= ReLU\left(\widetilde{F_{1}} \| (F_{2} \odot W)\right)
\end{equation}
where $(\cdot \| \cdot)$ means concatenating two feature maps along the channel dimension and $ReLU(\cdot)$ is ReLU activation function. 

\subsection{Texture Control}
The texture information of the face component region is difficult to capture due to its small area and diversity. To extract image-specific local texture, we define five face regions with rich textures: Left eye, right eye, left eyebrow, right eyebrow, and lip. For texture control, we explicitly extract the texture information of these regions from the reference image and inject them into the corresponding regions of the corrupted image.

% \noindent
% \textbf{Region-wise style code extraction}

\emph{1) Region-Wise Style Code Extraction: }
First, we obtain the segmentation map $S \in \mathbb{L}^{N \times H \times W}$ of the reference image via a well-trained face parsing network \cite{faceparsing}. Here, $N$ is the number of semantic classes defined by the parsing network. $H$ and $W$ represent the height and width. Then, we can find five pre-defined regions according to $S$, and distill their style codes from the reference image using the same style encoder of SEAN \cite{SEAN}. Specifically, a region-wise pooling layer is applied at the last layer of the style encoder, and we extract a 512-dimensional style code matrix $SM \in \mathbb{R}^{5 \times 512}$. Note that if a component has any pixels in the non-masked region, we ignore it and fill its style code value with 0. 

% \noindent
% \textbf{Component-Wise Style Injector (CWSI)}

\emph{2) Component-Wise Style Injector (CWSI):}
The structure of CWSI is shown in Fug. \ref{fig:framework}(c). Let $F \in \mathbb{R}^{C \times H \times W}$ denote the input feature maps. The generator must online predict a segmentation map associated with $F$ to locate the five pre-defined regions. To this end, we send $F$ into a segmentation branch to obtain its segmentation map $\bar{S} \in \mathbb{L}^{N \times H \times W}$ and select the five pre-defined regions $\widetilde{S} \in \mathbb{L}^{5 \times H \times W}$. The segmentation branch consists of two resnet blocks \cite{resnet} and a softmax layer. We broadcast the style code into the corresponding region according to $\widetilde{S}$ through a matrix multiplication:
\begin{equation}
z_{\text {style}}=SM^{\top} \times \widetilde{S}
\end{equation}
where $SM^{\top}$ denotes the transposed matrix of $SM$. Thus, $z_{\text {style}} \in \mathbb{R}^{512 \times H \times W}$ has the same layout as $\widetilde{S}$ but is filled by the style codes from $SM$. 
Finally, we use a denormalization operation similar to SPADE \cite{SPADE} to inject texture features. Specifically, we learn two parameters $\gamma$ and $\beta$ from $z_{\text {style}}$ to modulate $F$: 
\begin{equation}
\widetilde{F}= ReLU\left(\gamma \odot IN({F}) \oplus \beta\right)
\end{equation}
where $IN(\cdot)$ denotes instance normalization.

\begin{figure*}[t]
    \centering
    \includegraphics[width=17.0cm, trim=10 10 10 10,clip]{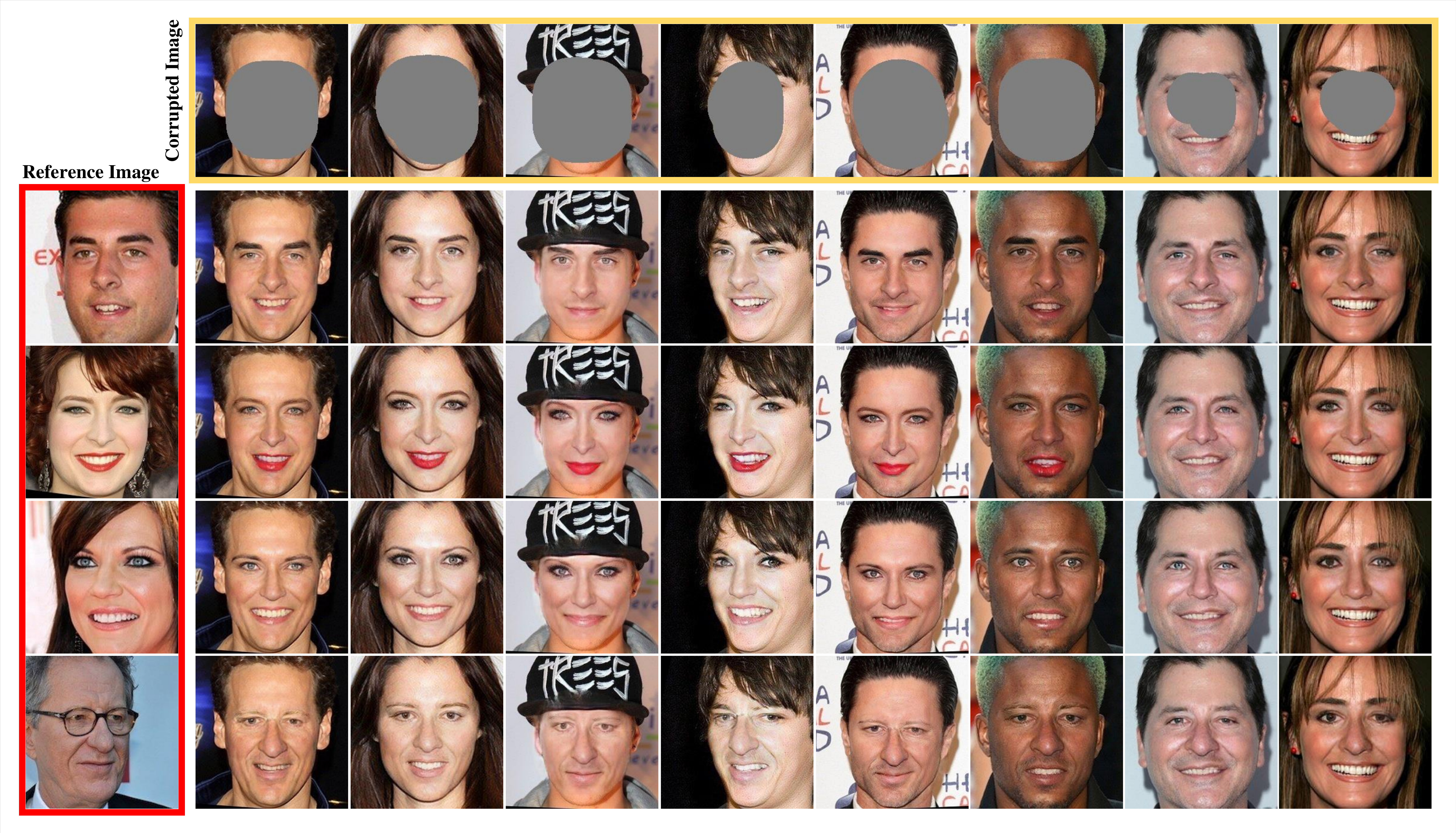}
    \caption{Inpainted results guided by different reference images.}
    \label{fig:cross_id}
\end{figure*}

\begin{figure}[t]
    \centering
    \includegraphics[width=8.5cm, trim=20 10 10 10,clip]{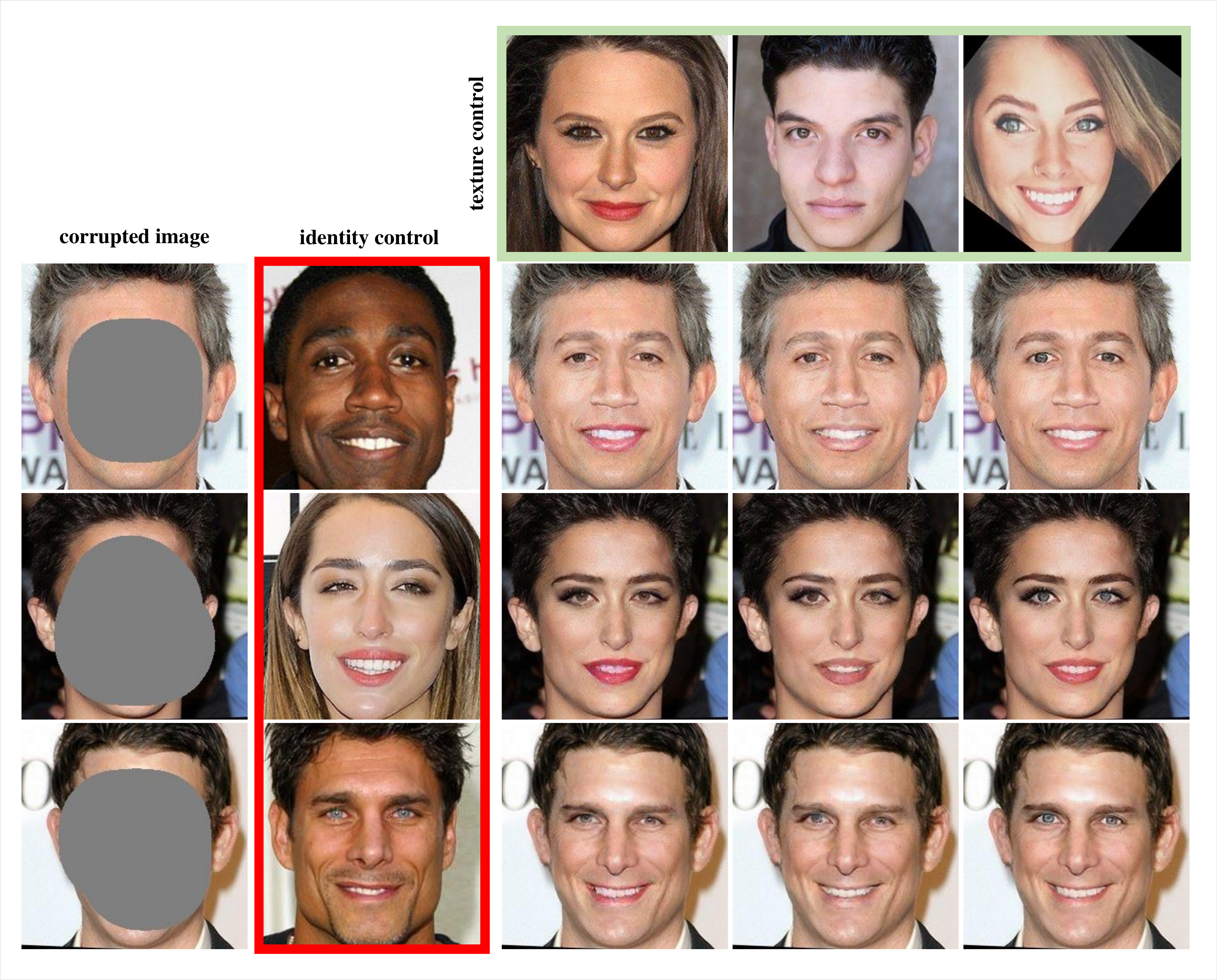}
    \caption{Control identity and texture using different reference images.}
    \label{fig:split_id_style}
\end{figure}

\begin{table*}[t]
\centering
\caption{Quantitative comparison with different mask rates. ($\uparrow$: Higher is better; $\downarrow$: Lower is better)}
\label{tab:sota}
% \scriptsize
% \scriptsize
% \setlength{\tabcolsep}{1.0mm}
\begin{tabular}{lccccccccc}
\Xhline{1pt}
\multicolumn{1}{c}{\multirow{2}{*}{Method}} & \multicolumn{3}{c}{Mask rate $\sim20$} & \multicolumn{3}{c}{Mask rate $\sim30$} & \multicolumn{3}{c}{Mask rate $\sim40$} \\ \cline{2-10} 
\multicolumn{1}{c}{}       & FID$\downarrow$        & LPIPS$\downarrow$     & IDR$\uparrow$       & FID$\downarrow$        & LPIPS$\downarrow$     & IDR$\uparrow$       & FID$\downarrow$        & LPIPS$\downarrow$     & IDR$\uparrow$       \\ \hline
ICT                     & 9.86       & 0.073     & 26.69     & 10.12      & 0.109     & 17.36     & 14.83      & 0.137     & 10.09      \\ \hline
CoMod                   & 7.29       & 0.068     & 23.81     & 9.01      & 0.089     & 20.50     & 11.81      & 0.128     & 9.32      \\ \hline
ICT+SimSwap             & 8.99      & 0.070     & 61.31     & 9.75      & 0.113     & 53.85      & 14.33      & 0.133     & 58.90      \\ \hline
CoMod+SimSwap           & 7.39      & 0.069     & 63.47     & 9.33      & 0.092     & 56.16     & 10.95      & 0.122     & 60.21     \\ \hline
Zhao et al.             & 36.66      & 0.123     & 59.87     & 36.33      & 0.188     & 56.16     & 35.57      & 0.222     & 56.51       \\ \hline
Ge et al.             & 10.03      & 0.083     & 29.15     & 10.34      & 0.98     & 21.91     & 12.15      & 0.129     & 11.21       \\ \hline
SwapInpaint             & 7.06      & 0.069     & 63.17     & 8.73      & 0.093     & 61.40     & 10.98      & 0.133     & 61.10       \\ \hline
Ours                    & \textbf{5.75}       & \textbf{0.059}     & \textbf{69.98}     &\textbf{7.67}         & \textbf{0.083}     & \textbf{66.85}     & \textbf{9.02}       & \textbf{0.116}     & \textbf{66.34}     \\ \bottomrule
\end{tabular}
\end{table*}

% \noindent
% \textbf{Training scheme for texture control}

\emph{3) Training Scheme For Texture Control: }
The success of texture control highly relies on the performance of CWSI's segmentation branch. If we cannot provide aligned supervision signals, texture control will fail, as described in Section 4.8. 
Ideally, we need an aligned segmentation map $S_{f}$ to match the current inpainted result online when training, as illustrated in Fig. \ref{fig:training_mode}(a). 
In practice, we can only pre-generate $S_{gt}$ from ground truth as supervision signals for segmentation branches. However, it cannot be guaranteed that $S_{gt}$ could be aligned with the inpainted result precisely in the working pipeline when the masked region is large, especially for small eye and mouth regions.
The general training scheme suffers from always appearing misalignment, if we only utilizes regular masks (\emph{inpainting mode}) to generate corrupted images, where the training scheme as such is referred to as Single-Mode Training (SMT) depicted in Fig. \ref{fig:training_mode}(c). 
To overcome the misalignment problem, we introduce a Three-Mode Training scheme (TMT) with two additional modes referred to as \emph{segmentation mode} and \emph{style extracting mode}, as shown in Fig. \ref{fig:training_mode}(b).
\emph{Segmentation mode} utilizes a small mask that does not cover the face’s central region and takes ground truth as the reference image. 
Under such settings, we force the generator to focus on improving the performance of the segmentation branch.
\emph{Style extracting mode} employs a special mask corresponding with the five pre-defined facial components to enhance training on the facial components for texture control injection. 
In two additional modes, $S_{gt}$ is aligned with the inpainted result because missing regions are set to be small.
TMT applies three training modes iteratively for each training step. Specifically, we alternately perform \emph{segmentation mode} and \emph{style extracting mode} between two adjacent \emph{inpainting modes}, as described in Fig. \ref{fig:training_mode}(d).

\subsection{Loss Function}

% \noindent
% \textbf{Pixel Reconstruction Loss} $\mathcal{L}_{R}$

\emph{1) Pixel Reconstruction Loss} $\mathcal{L}_{R}$  
calculates the L1 distance between the inpainted image and the ground truth in RGB space.

% \noindent
% \textbf{Perceptual Loss}  $\mathcal{L}_{P}$

\emph{2) Perceptual Loss} $\mathcal{L}_{P}$  
\cite{johnson2016perceptual} is used to force the output of the generator to be closer to the ground truth in a pre-trained VGG-19 \cite{vgg} feature space at multiple layers. 

% \noindent
% \textbf{Identity-preserving Loss} $\mathcal{L}_{id}$

\emph{3) Identity-preserving Loss} $\mathcal{L}_{id}$  
is used to force the generated image's identity to be consistent with the reference image's identity. It is formulated as:
\begin{equation}
\mathcal{L}_{id}=1- \boldsymbol\cos \left(\boldsymbol{z}_{id}\left(I_{G}\right), \boldsymbol{z}_{i d}\left(I_{r}\right)\right)
\end{equation}
where $\cos (\cdot, \cdot)$ represents the cosine similarity of two vectors.

% \noindent
% \textbf{Semantic Segmentation Loss} $\mathcal{L}_{S}$ 

\emph{4) Semantic Segmentation Loss} $\mathcal{L}_{S}$  
is employed for training the segmentation branches of CWSI. We supervise the output of each segmentation branch using cross-entropy loss of semantic segmentation task \cite{FCN}.

% \noindent
% \textbf{Global Adversarial Loss} $\mathcal{L}_{adv,G}$ 

\emph{5) Global Adversarial Loss} $\mathcal{L}_{adv,G}$ 
follows SNPatchGAN \cite{GC} but with the hinge version \cite{brock2018large, liu2019few}. It is applied to the entire image and evaluates whether the entire image is coherent as a whole.

% \noindent
% \textbf{Local Adversarial Loss} $\mathcal{L}_{adv,L}$

\emph{6) Local Adversarial Loss} $\mathcal{L}_{adv,L}$ 
is calculated on three sub-regions (left eye, right eye, and mouth) using the same loss function as the global adversarial loss following the previous work \cite{zhou2020learning}. Then, the losses of the three regions are summed as the final local adversarial loss. 

The overall Loss can be written as:
\begin{equation}
\mathcal{L}= \lambda_{R} \mathcal{L}_{R} + \lambda_{P} \mathcal{L}_{\mathrm{P}} + \lambda_{id} \mathcal{L}_{id} + \lambda_{S} \mathcal{L}_{S} + \mathcal{L}_{adv, G} + I_{adv, L} \mathcal{L}_{adv, L}
\end{equation}
Here, $\lambda_{R}=20$, $\lambda_{P}=10$, $\lambda_{id}=3$, $\lambda_{S}=2.5$. $I_{adv, L}$ is 1 in \emph{inpainting mode} and 0 in other modes.

\section{Experiments}
\subsection{Dataset}
Our task needs a high-quality dataset for training. Based on Celeb-ID \cite{eye}, we build a dataset suitable for our task.  Celeb-ID contains around 1700 individual identities and a total of 100K images, but it contains a large number of extremely blurred and occluded photos. We manually eliminated the low-quality images. In the end, our dataset contains a total of 16,259 identities and 94,770 images and at least two photos are provided for each identity. 1,500 identities are used for testing, and the rest are used for training. 

\subsection{Evaluation Metrics}
% \noindent
% \textbf{Fr\'echet Inception Distance (FID)}

\emph{1) Fr\'echet Inception Distance (FID)}
\cite{FID} utilizes a pre-trained Inception-v3 \cite{szegedy2016rethinking} network to extract features and measure the distance between generated samples and real samples. It has been widely demonstrated that it is consistent with human visual perception. Lower FID value indicates higher fidelity.

% \noindent
% \textbf{Learned perceptual image patch similarity (LPIPS)}

\emph{2) Learned perceptual image patch similarity (LPIPS)}
\cite{LPIPS} evaluates the similarity between the generated image and the corresponding ground truth in a pairwise manner. A lower LPIPS indicates that the generated image is closer to the target.

% \noindent
% \textbf{ID retrieval accuracy (IDR)}

\emph{3) ID retrieval accuracy (IDR): }
We randomly select two images as the corrupted  images and the reference image for each identity in the test set, respectively. The unselected images are put into a retrieval pool. The identity embeddings are extracted from the inpainted image and the retrieval pool images, using a different face recognition model \cite{wang2018cosface}. Then, the pair-wise distance is calculated using cosine similarity. We look for the nearest sample in the retrieval pool for each completed result and check whether they belong to the same identity. The accuracy of such retrieval is reported as identity retrieval accuracy (IDR).

\subsection{Implementation Details}
All images are resized to $256 \times 256$, and the cropped sizes of eyes and mouth are $55 \times 60$, and $44 \times 90$, respectively, in calculating the local adversarial loss. In order to simulate a more general application scenario, we use free-form masks with different sizes. We employ Adam \cite{adam} optimizers for both the generator and the discriminators with momentum $\beta_{1}=0.5$ and $\beta_{2}=0.999$. The learning rate is set to 0.0002. We only update the parameters of the style encoder in \emph{style extracting mode}. The proposed model is trained for 300 epochs on a single RTX 3090 GPU.

\subsection{Baselines}
We introduce additional three types of baselines for an extensive comparison.

% \noindent
% \textbf{Reference-guided face inpainting} 

\emph{1) Reference-Guided Face Inpainting Method: }
Zhao et al. \cite{BMVC,tcsvt,li2021swapinpaint} are the most relevant works to our method and also uses a reference image to control inpainted results.

% \noindent
% \textbf{General image inpainting} 

\emph{2) General Image Inpainting Method: }
We investigate the performance of two currently leading inpainting methods, ICT \cite{wan2021high} and CoMod \cite{comod}. They show impressive results dealing with large-region masks. 

% \noindent
% \textbf{Combination of image inpainting and face swapping} 

\emph{3) Combination Of Image Inpainting And Face Swapping:}
Because the general inpainting method does not employ the reference image, we combine the image inpainting and face swapping as new baselines for a fairer comparison. Specifically, the results of ICT or CoMod with reference images are fed into a face-swapping method SimSwap \cite{chen2020simswap}. SimSwap is capable of transferring the identity of an arbitrary source face into an arbitrary target face. Some famous open-source tools \cite{faceswap,DeepFaceLab,nirkin2019fsgan} are not considered because they require specialized training for different inputs. 

\begin{table*}[t]
\centering
\caption{Quantitative comparison of ablation studies.}
\label{tab:ablation}
% \scriptsize
% \scriptsize
% \setlength{\tabcolsep}{0.8mm}
\begin{tabular}{lcccccccccccc}
\Xhline{1pt}
\multicolumn{1}{c}{\multirow{2}{*}{Method}} & \multicolumn{4}{c}{Mask rate $\sim20$} & \multicolumn{4}{c}{Mask rate $\sim30$} & \multicolumn{4}{c}{Mask rate $\sim40$} \\ \cline{2-13} 
\multicolumn{1}{c}{}       & FID$\downarrow$       & LPIPS$\downarrow$      & IDR$\uparrow$     & FRR$\uparrow$      & FID$\downarrow$       & LPIPS$\downarrow$      & IDR$\uparrow$   & FRR$\uparrow$       & FID$\downarrow$        & LPIPS$\downarrow$     & IDR$\uparrow$      & FRR$\uparrow$      \\ \hline
\emph{w AdaIN}                    & 7.93      & 0.081       & 65.13     & 48.80    & 10.04      & 0.094      & 60.25     &  46.13    & 11.16     & 0.130     & 62.18     &  47.33   \\ \hline
\emph{w/o Half-AdaIN}             & 7.04      & 0.062      & 47.75     & 11.33     & 9.79      & 0.095      & 27.57     &  5.93    & 11.49       & 0.143     & 23.90     & 1.40    \\ \hline
\emph{w/o CWSI}                   & 6.04      & 0.060      & 69.12     & 61.33     & 8.30      & 0.090      & \textbf{69.31}     & 59.80    & 10.29      & 0.118     & 65.93     & 57.33    \\ \hline
\emph{w SMT}                      & 6.73      & 0.069      & 68.03     &  60.67    & 8.73      & 0.091      & 64.33     &  60.27   & 11.08      & 0.129     & 63.79     & 58.26    \\ \hline
Full model (\emph{w TMT})                 & \textbf{5.75}       & \textbf{0.059}     & \textbf{69.98}     & \textbf{67.60}     &\textbf{7.67}         & \textbf{0.083}     & 66.85     & \textbf{68.07}     & \textbf{9.02}       & \textbf{0.116}     & \textbf{66.34}     & \textbf{67.87}    \\ \bottomrule
\end{tabular}
\end{table*}

\begin{table}[t]
\centering
\caption{Comparison of segmentation performance with different training schemes.}
\label{tab:TMM}
\setlength{\tabcolsep}{1.0mm}
\begin{tabular}{lcc}
\Xhline{1pt}
                 & mIoU$\uparrow$           & Acc$\uparrow$ \\ \hline
\textit{w SMT} & 0.543          & 0.898          \\ \hline
\textit{w TMT}   & \textbf{0.698} & \textbf{0.947} \\ \bottomrule
\end{tabular}
\end{table}

\begin{figure}[t]
    \centering
    \includegraphics[width=8.5cm, trim=10 10 10 10,clip]{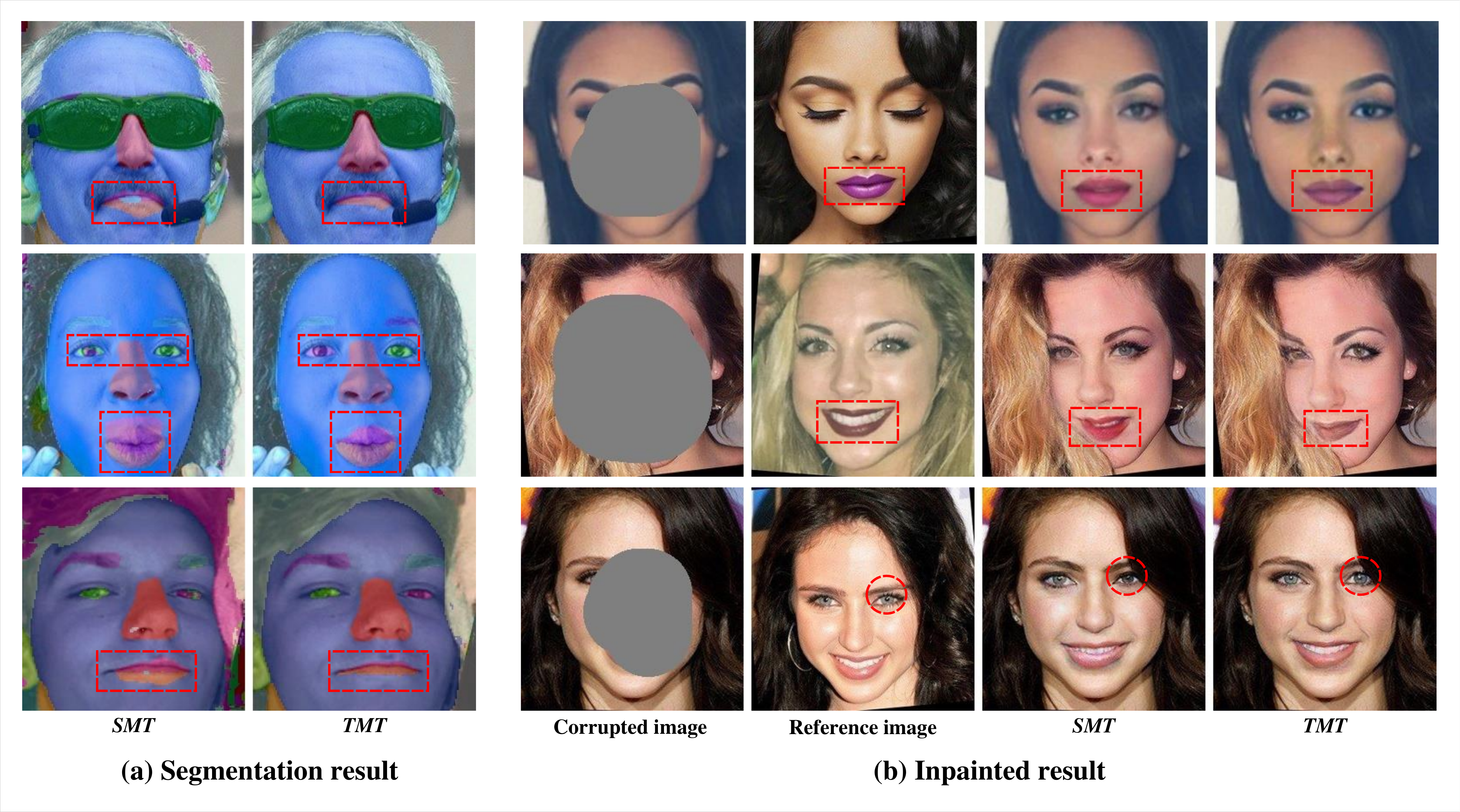}
    \caption{(a) Visualize segmentation maps predicted by the segmentation branch. Different colors represent different semantic classes. (b) SMT causes color inconsistency between the inpainted results and the reference images.}
    \label{fig:SMT_TMT}
\end{figure}

\begin{figure}[t]
    \centering
    \includegraphics[width=9.0cm, trim=25 10 130 10,clip]{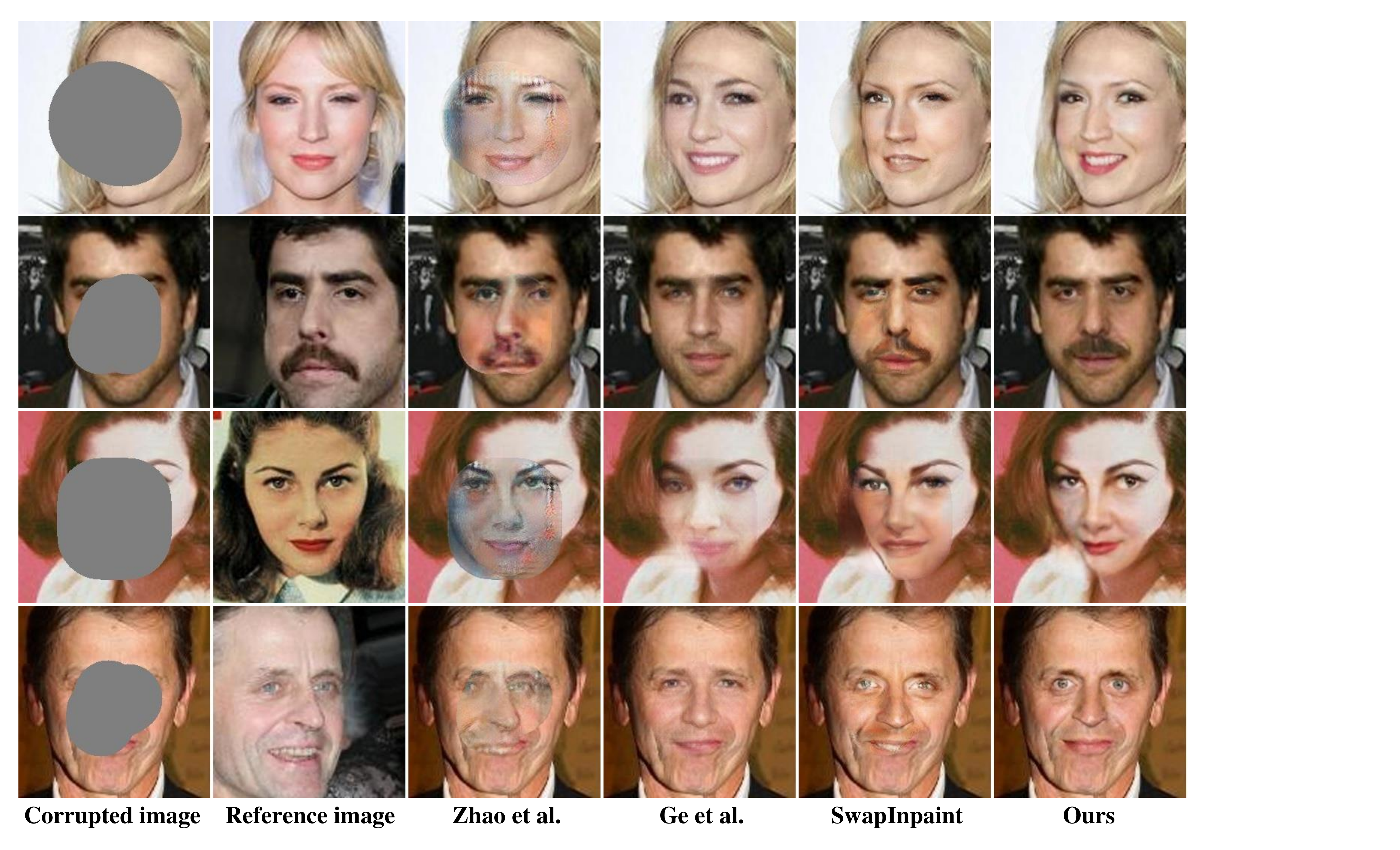}
    \caption{Visual comparison on CelebA dataset.}
    \label{fig:sota_celebA}
\end{figure}

\begin{figure}[t]
    \centering
    \includegraphics[width=8.5cm, trim=25 10 10 10,clip]{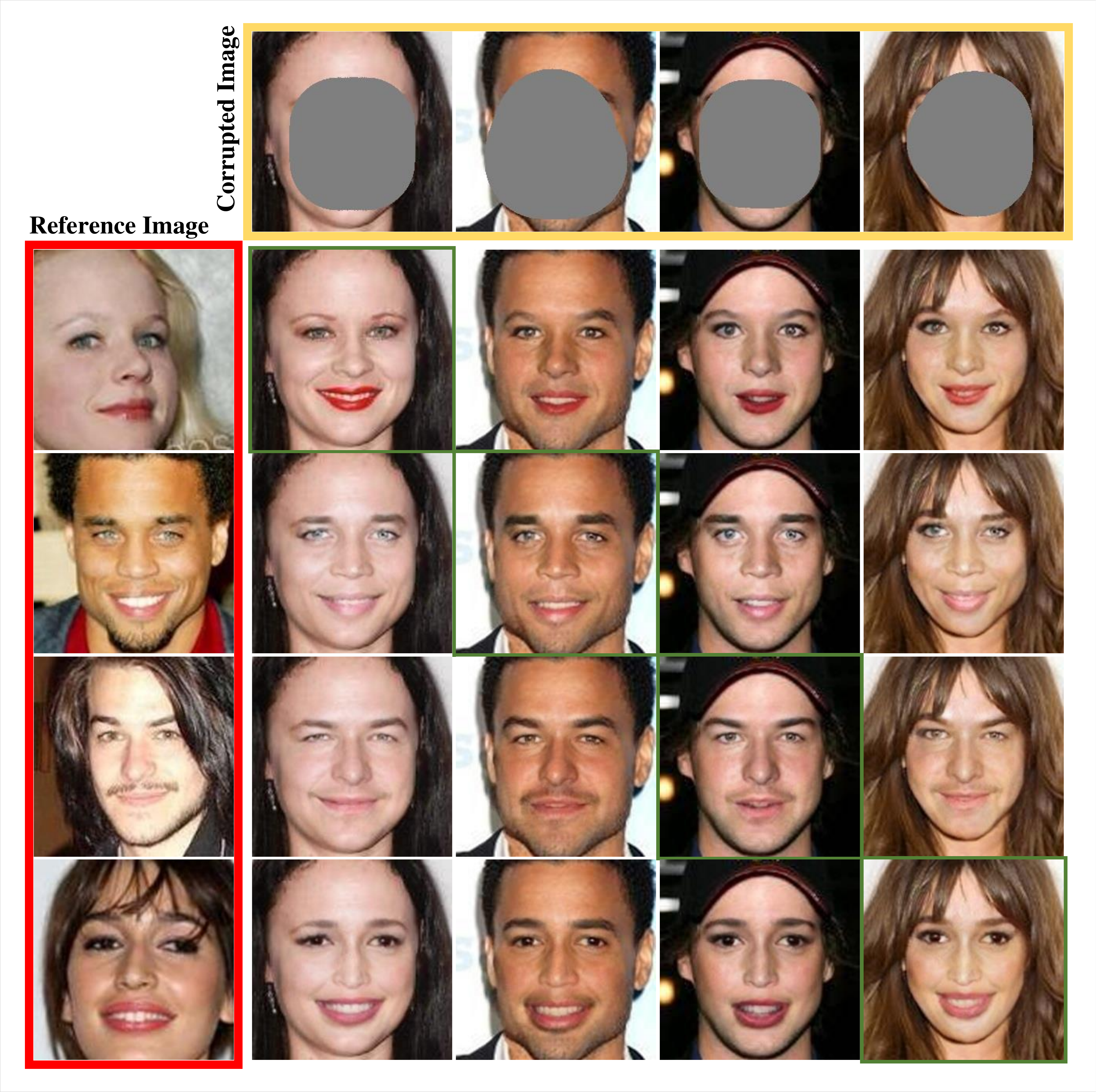}
    \caption{Cross-identity results on CelebA dataset.}
    \label{fig:cross_id_celebA}
\end{figure}

\begin{table}[t]
\centering
\caption{Quantitative Comparison On CelebA Dataset.}
\label{tab:sota_celebA}
\setlength{\tabcolsep}{1.0mm}
\begin{tabular}{lccc}
\Xhline{1pt}
\multicolumn{1}{c}{Method} & FID$\downarrow$       & LPIPS$\downarrow$      & IDR$\uparrow$   \\ \hline
Zhao et al.                & 123.12 & 0.497 & 39.25 \\ \hline
Ge et al.                  & 80.10  & 0.319 & 13.10 \\ \hline
SwapInpaint                & 79.12  & 0.309 & 43.70 \\ \hline
Ours                       & \textbf{61.34}  & \textbf{0.237} & \textbf{53.75} \\ \bottomrule
\end{tabular}
\end{table}

\begin{figure}[t]
    \centering
    \includegraphics[width=6.0cm, trim=25 10 10 10,clip]{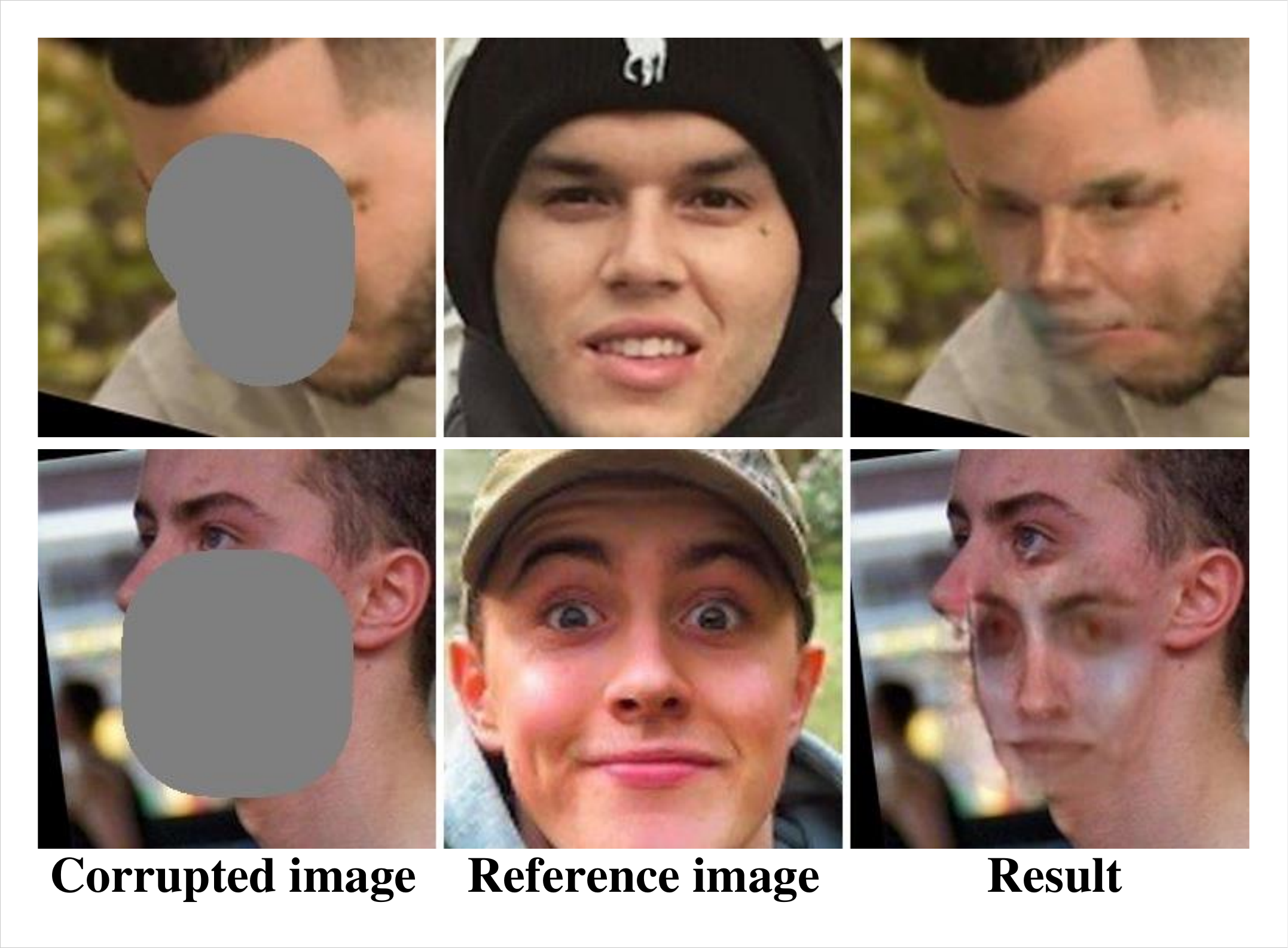}
    \caption{Failure cases.}
    \label{fig:failure_cases}
\end{figure}

\begin{table}[h]
\centering
% \scriptsize
\caption{Comparisons of inference speed and FLOPs.}
\setlength{\tabcolsep}{1.0mm}
\begin{tabular}{lcccccc}
\Xhline{1pt}
                & Running time (Sec.)       & FLOPs (MB) \\ \hline
Zhao et al.     & 0.0070                    & 3045      \\ \hline
Ge et al.       & 0.0109                    & 3463       \\ \hline
SwapInpaint     & 0.0453                    & 5733       \\ \hline
Ours            & 0.0372                    & 4005       \\ \bottomrule
\end{tabular}
\label{tab:efficiency2}
\end{table}

\subsection{Quantitative and Qualitative Comparison}
Fig. \ref{fig:sota} provides a visual comparison. Zhao et al. \cite{BMVC} suffers from serious style inconsistency and misalignment. Since \cite{BMVC} directly uses the reference image as the condition, it is difficult for their generator to decompose information into identity and texture from the reference image.
General face inpainting \cite{wan2021high,comod} methods are able to produce visually pleasing results. However, they lack controllability and cannot produce desired results guided by specific reference images. Combination methods can control the inpainted result, but they suffer from several significant drawbacks:
(1) Simply combining two methods leads to error accumulation. For example, face swapping will produce unpleasant results if the inpainting method produces some distorted structures. (2) Although the face swapping approach can inject identity information, it ignores the texture information provided by the reference image.
Ge et al. \cite{tcsvt} only employ identity loss to constrain the inpainted results, which cannot effectively preserve identity information. SwapInpaint \cite{li2021swapinpaint} can generate identity-consistent results but ignores local texture information provided by the reference image.
Our method can generate high-quality results consistent with the reference image in terms of identity and texture. Significant performance gains benefit from our separate injection of identity and texture information, as well as the end-to-end joint optimization in the proposed comprehensive framework. As shown in Table \ref{tab:sota}, the quantitative results also demonstrate that our method can yield realistic results and maintain a high degree of consistency with reference images.

\subsection{Cross-Identity Face Inpainting}
Our model can also achieve cross-identity face inpainting, which is similar to face swapping.  However, it should be noted that our task is fundamentally different from face swapping. Face swapping can see the layout of the source image, but our task needs to predict the content in the missing region. Correspondingly, our results are demonstrated in Fig. \ref{fig:cross_id}. 

\subsection{Separately Control Identity and Texture}
Due to our decoupled design, our model can flexibly control identity and texture. Furthermore, we use two different reference images to generate identity vectors and style codes to demonstrate how the decoupled controls work in cooperation with each other. Some visual results are shown in Fig. \ref{fig:split_id_style}. 

\subsection{Ablation Study}
% \noindent
% \textbf{Identity Control}

\emph{1) Identity Control: }
Identity information is introduced to control the identity of the result. We replace all Half-AdaIN with standard convolutional layers but still use identity preserving loss (\emph{w AdaIN}). As shown in Fig. \ref{fig:ablation}, the model without identity control cannot recall the identity information of the reference image. The quantitative results in Table \ref{tab:ablation} show the effectiveness of identity control, especially from the perspective of retrieval accuracy. 
We also report \emph{Face Recognition Rate (FRR)}. Specifically, we utilize a well-trained face recognizer \cite{deng2019arcface} to calculate the distance between generated and reference images. The threshold value is set to 0.7

% \noindent
% \textbf{Effect of Half-AdaIN}

\emph{2) Effect of Half-AdaIN: }
We design Half-AdaIN to inject identity information for generating high-quality results. The distinct point of Half-AdaIN is that it preserves the contextual style from corrupted images.
We replace all Half-AdaIN with AdaIN (\emph{w/o Half-AdaIN}). The visual results drop significantly and produce style inconsistency as illustrated in \ref{fig:ablation}. Quantitative results are given in Table \ref{tab:ablation}, which indicates that Half-AdaIN contributes to the performance gain. 

% \noindent
% \textbf{Effect of CWSI}

\emph{3) Effect of CWSI: }
We develop CWSI to effectively control the texture of face component regions. We remove all CWSI in our model as a baseline (\emph{w/o CWSI}). Fig. \ref{fig:ablation} shows that the results generated by baseline are inconsistent with the reference image in texture style. This shows that CWSI can effectively extract the texture information of pre-defined regions from the reference image and inject them into the corresponding regions of the inpainted image. 

% \noindent
% \textbf{Effect of Training Scheme}

\emph{4) Effect of Training Scheme:}
Three-mode Training (\emph{TMT}) scheme is designed for more efficient generation of intermediate segmentation maps and accurate injection of texture features into corresponding regions. We set a baseline (\emph{"w SMT"}) using Single-Mode Training scheme.
We apply the masks similar to those of \emph{segmentation mode}
, so the outputs of the segmentation branch should be the same as ground truth segmentation maps. Then, we evaluate the performance of the segmentation branch with the resolution of $128 \times 128$ for the two different training schemes. 
Some visual results are shown in Fig. \ref{fig:SMT_TMT}(a). It is clear that \emph{SMT} produces more false predictions on small face component regions, especially in mouth and eye regions. Quantitative results in Table \ref{tab:TMM} show that \emph{TMT} helps the generator significantly improve the segmentation branch's performance in terms of \emph{Mean Intersection over Union} (mIoU) and \emph{Pixel Accuracy} (Acc). 
We demonstrate some inpainted examples, as shown in Fig. \ref{fig:SMT_TMT}(b). We observe an apparent texture inconsistency between \emph{SMT}'s results and the reference images in component regions. This inconsistency is caused by \emph{SMT} not generating accurate segmentation maps to locate face components.

\subsection{Cross-Dataset Evaluation}
To demonstrate the generalization of the proposed method, we conduct a cross-dataset evaluation on CelebA dataset \cite{liu2015faceattributes}. The generator is trained with the Celeb-ID dataset, where we utilize a pre-trained arcface model \cite{InsightFace-v2} to extract identity vectors.
We randomly selected 2000 identities from CelebA dataset and generated masks with random rates. The Qualitative and quantitative comparisons, as shown in Fig.  \ref{fig:sota_celebA} and Table \ref{tab:sota_celebA}, reveal that the proposed method can generate more realistic results and achieve better consistency with reference images.
The results in Fig. \ref{fig:cross_id_celebA} demonstrate more results of our model on CelebA dataset.

\subsection{Failure cases}
We show some failure cases in Figure \ref{fig:failure_cases}. Our model may produce unsatisfactory results if the input images have extreme poses. There are two reasons for this phenomenon: (1) Most training set images are frontal. Images with extreme poses are rare. Therefore, the generator does not have enough training samples. (2) Face images with extreme poses are difficult to inpaint due to their diversity and complexity.

\subsection{Efficiency}
The average running time per image and FLOPs during testing are listed in Table \ref{tab:efficiency2}. This experiment is performed on a NVIDIA RTX 3090 GPU.

\section{Conclusion}
In this paper, we propose a comprehensive framework for reference-guided face inpainting. Our approach can efficiently control the generated results guided by a reference image. To accurately inject two types of control information and produce high-quality results, we gracefully designed modules: Half-AdaIN and CWSI. Half-AdaIN is a variant of AdaIN, which injects identity information while preserving the contextual style of the input face. CWSI can inject component-specific texture information precisely into the corresponding face regions by parsing feature maps. The proposed model can provide separate control for identity and texture over missing regions. Extensive experiments verify the superiority and practicability of our method. However, our method remains limited in processing face images with extreme poses, which needs to be addressed in future work.

% \section*{Acknowledgment}

% %Dr. Reveryrand would like to acknowledge the funding by XLIM, Limoges, France. 
% The authors would like to thank Dr. David Root and Dr. Jean-Pierre Teyssier at Agilent Technologies for the loan of the time-domain nonlinear measurement equipment and TriQuint Semiconductor for the donation of the transistors. 

% or


% ============================================
%\appendices
%Appendix one text goes here %\cite{Roberg2010}.

%\section{}

%\section*{Acknowledgment}

%The authors would like to thank D. Root for the loan of the SWAP. The SWAP that can ONLY be usefull in Boulder...

\ifCLASSOPTIONcaptionsoff
  \newpage
\fi



%\begin{thebibliography}{1}


\bibliographystyle{IEEEtran}
\bibliography{IEEEabrv,Bibliography}

\begin{thebibliography}{10}
\providecommand{\url}[1]{#1}
\csname url@rmstyle\endcsname
\providecommand{\newblock}{\relax}
\providecommand{\bibinfo}[2]{#2}
\providecommand\BIBentrySTDinterwordspacing{\spaceskip=0pt\relax}
\providecommand\BIBentryALTinterwordstretchfactor{4}
\providecommand\BIBentryALTinterwordspacing{\spaceskip=\fontdimen2\font plus
\BIBentryALTinterwordstretchfactor\fontdimen3\font minus
  \fontdimen4\font\relax}
\providecommand\BIBforeignlanguage[2]{{%
\expandafter\ifx\csname l@#1\endcsname\relax
\typeout{** WARNING: IEEEtran.bst: No hyphenation pattern has been}%
\typeout{** loaded for the language `#1'. Using the pattern for}%
\typeout{** the default language instead.}%
\else
\language=\csname l@#1\endcsname
\fi
#2}}

\bibitem{duan2021simultaneous}
Q.~Duan, L.~Zhang, and X.~Gao, ``Simultaneous face completion and
  frontalization via mask guided two-stage gan,'' \emph{IEEE Transactions on
  Circuits and Systems for Video Technology}, 2021.

\bibitem{tu2021joint}
X.~Tu, J.~Zhao, Q.~Liu, W.~Ai, G.~Guo, Z.~Li, W.~Liu, and J.~Feng, ``Joint face
  image restoration and frontalization for recognition,'' \emph{IEEE
  Transactions on Circuits and Systems for Video Technology}, vol.~32, no.~3,
  pp. 1285--1298, 2021.

\bibitem{lahiri2020lightweight}
A.~Lahiri, S.~Bairagya, S.~Bera, S.~Haldar, and P.~K. Biswas, ``Lightweight
  modules for efficient deep learning based image restoration,'' \emph{IEEE
  Transactions on Circuits and Systems for Video Technology}, vol.~31, no.~4,
  pp. 1395--1410, 2020.

\bibitem{xia2021local}
Y.~Xia, W.~Zheng, Y.~Wang, H.~Yu, J.~Dong, and F.-Y. Wang, ``Local and global
  perception generative adversarial network for facial expression synthesis,''
  \emph{IEEE Transactions on Circuits and Systems for Video Technology},
  vol.~32, no.~3, pp. 1443--1452, 2021.

\bibitem{tcsvt}
S.~Ge, C.~Li, S.~Zhao, and D.~Zeng, ``Occluded face recognition in the wild by
  identity-diversity inpainting,'' \emph{IEEE Transactions on Circuits and
  Systems for Video Technology}, vol.~30, no.~10, pp. 3387--3397, 2020.

\bibitem{BMVC}
Y.~Zhao, W.~Chen, J.~Xing, X.~Li, Z.~Bessinger, F.~Liu, W.~Zuo, and R.~Yang,
  ``Identity preserving face completion for large ocular region occlusion.''

\bibitem{li2021swapinpaint}
H.~Li, W.~Wang, C.~Yu, and S.~Zhang, ``Swapinpaint: Identity-specific face
  inpainting with identity swapping,'' \emph{IEEE Transactions on Circuits and
  Systems for Video Technology}, 2021.

\bibitem{Adain}
X.~Huang and S.~Belongie, ``Arbitrary style transfer in real-time with adaptive
  instance normalization,'' in \emph{Proceedings of the IEEE International
  Conference on Computer Vision}, 2017, pp. 1501--1510.

\bibitem{chen2020simswap}
R.~Chen, X.~Chen, B.~Ni, and Y.~Ge, ``Simswap: An efficient framework for high
  fidelity face swapping,'' in \emph{Proceedings of the 28th ACM International
  Conference on Multimedia}, 2020, pp. 2003--2011.

\bibitem{luo2022context}
W.~Luo, S.~Yang, H.~Wang, B.~Long, and W.~Zhang, ``Context-consistent semantic
  image editing with style-preserved modulation,'' in \emph{European Conference
  on Computer Vision}.\hskip 1em plus 0.5em minus 0.4em\relax Springer, 2022,
  pp. 561--578.

\bibitem{ballester2001filling}
C.~Ballester, M.~Bertalmio, V.~Caselles, G.~Sapiro, and J.~Verdera,
  ``Filling-in by joint interpolation of vector fields and gray levels,''
  \emph{IEEE transactions on image processing}, vol.~10, no.~8, pp. 1200--1211,
  2001.

\bibitem{bertalmio2000image}
M.~Bertalmio, G.~Sapiro, V.~Caselles, and C.~Ballester, ``Image inpainting,''
  in \emph{Proceedings of the 27th annual conference on Computer graphics and
  interactive techniques}, 2000, pp. 417--424.

\bibitem{levin2003learning}
A.~Levin, A.~Zomet, and Y.~Weiss, ``Learning how to inpaint from global image
  statistics,'' in \emph{Proceedings of the Ninth IEEE International Conference
  on Computer Vision-Volume 2}, 2003, p. 305.

\bibitem{efros1999texture}
A.~A. Efros and T.~K. Leung, ``Texture synthesis by non-parametric sampling,''
  in \emph{Proceedings of the seventh IEEE international conference on computer
  vision}, vol.~2.\hskip 1em plus 0.5em minus 0.4em\relax IEEE, 1999, pp.
  1033--1038.

\bibitem{kwatra2005texture}
V.~Kwatra, I.~Essa, A.~Bobick, and N.~Kwatra, ``Texture optimization for
  example-based synthesis,'' in \emph{ACM SIGGRAPH 2005 Papers}, 2005, pp.
  795--802.

\bibitem{efros2001image}
A.~A. Efros and W.~T. Freeman, ``Image quilting for texture synthesis and
  transfer,'' in \emph{Proceedings of the 28th annual conference on Computer
  graphics and interactive techniques}, 2001, pp. 341--346.

\bibitem{GAN}
I.~Goodfellow, J.~Pouget-Abadie, M.~Mirza, B.~Xu, D.~Warde-Farley, S.~Ozair,
  A.~Courville, and Y.~Bengio, ``Generative adversarial nets,'' in
  \emph{Advances in neural information processing systems}, 2014, pp.
  2672--2680.

\bibitem{pathak2016context}
D.~Pathak, P.~Krahenbuhl, J.~Donahue, T.~Darrell, and A.~A. Efros, ``Context
  encoders: Feature learning by inpainting,'' in \emph{Proceedings of the IEEE
  conference on computer vision and pattern recognition}, 2016, pp. 2536--2544.

\bibitem{iizuka2017globally}
S.~Iizuka, E.~Simo-Serra, and H.~Ishikawa, ``Globally and locally consistent
  image completion,'' \emph{ACM Transactions on Graphics (ToG)}, vol.~36,
  no.~4, pp. 1--14, 2017.

\bibitem{CA}
J.~Yu, Z.~Lin, J.~Yang, X.~Shen, X.~Lu, and T.~S. Huang, ``Generative image
  inpainting with contextual attention,'' in \emph{Proceedings of the IEEE
  conference on computer vision and pattern recognition}, 2018, pp. 5505--5514.

\bibitem{zeng2019learning}
Y.~Zeng, J.~Fu, H.~Chao, and B.~Guo, ``Learning pyramid-context encoder network
  for high-quality image inpainting,'' in \emph{Proceedings of the IEEE/CVF
  Conference on Computer Vision and Pattern Recognition}, 2019, pp. 1486--1494.

\bibitem{zeng2021cr}
Y.~Zeng, Z.~Lin, H.~Lu, and V.~M. Patel, ``Cr-fill: Generative image inpainting
  with auxiliary contextual reconstruction,'' in \emph{Proceedings of the
  IEEE/CVF International Conference on Computer Vision}, 2021, pp.
  14\,164--14\,173.

\bibitem{liu2022deep}
J.~Liu, M.~Gong, Z.~Tang, A.~Qin, H.~Li, and F.~Jiang, ``Deep image inpainting
  with enhanced normalization and contextual attention,'' \emph{IEEE
  Transactions on Circuits and Systems for Video Technology}, 2022.

\bibitem{wang2022dual}
C.~Wang, M.~Shao, D.~Meng, and W.~Zuo, ``Dual-pyramidal image inpainting with
  dynamic normalization,'' \emph{IEEE Transactions on Circuits and Systems for
  Video Technology}, 2022.

\bibitem{GC}
J.~Yu, Z.~Lin, J.~Yang, X.~Shen, X.~Lu, and T.~S. Huang, ``Free-form image
  inpainting with gated convolution,'' in \emph{Proceedings of the IEEE
  International Conference on Computer Vision}, 2019, pp. 4471--4480.

\bibitem{pconv}
G.~Liu, F.~A. Reda, K.~J. Shih, T.-C. Wang, A.~Tao, and B.~Catanzaro, ``Image
  inpainting for irregular holes using partial convolutions,'' in
  \emph{Proceedings of the European Conference on Computer Vision (ECCV)},
  2018, pp. 85--100.

\bibitem{nazeri2019edgeconnect}
K.~Nazeri, E.~Ng, T.~Joseph, F.~Qureshi, and M.~Ebrahimi, ``Edgeconnect:
  Structure guided image inpainting using edge prediction,'' in
  \emph{Proceedings of the IEEE/CVF International Conference on Computer Vision
  Workshops}, 2019, pp. 0--0.

\bibitem{ren2019structureflow}
Y.~Ren, X.~Yu, R.~Zhang, T.~H. Li, S.~Liu, and G.~Li, ``Structureflow: Image
  inpainting via structure-aware appearance flow,'' in \emph{Proceedings of the
  IEEE/CVF International Conference on Computer Vision}, 2019, pp. 181--190.

\bibitem{xu2020e2i}
S.~Xu, D.~Liu, and Z.~Xiong, ``E2i: Generative inpainting from edge to image,''
  \emph{IEEE Transactions on Circuits and Systems for Video Technology},
  vol.~31, no.~4, pp. 1308--1322, 2020.

\bibitem{wang2020structure}
C.~Wang, X.~Chen, S.~Min, J.~Wang, and Z.-J. Zha, ``Structure-guided deep video
  inpainting,'' \emph{IEEE Transactions on Circuits and Systems for Video
  Technology}, vol.~31, no.~8, pp. 2953--2965, 2020.

\bibitem{zheng2019pluralistic}
C.~Zheng, T.-J. Cham, and J.~Cai, ``Pluralistic image completion,'' in
  \emph{Proceedings of the IEEE/CVF Conference on Computer Vision and Pattern
  Recognition}, 2019, pp. 1438--1447.

\bibitem{liu2021pd}
H.~Liu, Z.~Wan, W.~Huang, Y.~Song, X.~Han, and J.~Liao, ``Pd-gan: Probabilistic
  diverse gan for image inpainting,'' in \emph{Proceedings of the IEEE/CVF
  Conference on Computer Vision and Pattern Recognition}, 2021, pp. 9371--9381.

\bibitem{feng2022generative}
X.~Feng, W.~Pei, F.~Li, F.~Chen, D.~Zhang, and G.~Lu, ``Generative
  memory-guided semantic reasoning model for image inpainting,'' \emph{IEEE
  Transactions on Circuits and Systems for Video Technology}, 2022.

\bibitem{wan2021high}
Z.~Wan, J.~Zhang, D.~Chen, and J.~Liao, ``High-fidelity pluralistic image
  completion with transformers,'' in \emph{Proceedings of the IEEE/CVF
  International Conference on Computer Vision}, 2021, pp. 4692--4701.

\bibitem{li2017generative}
Y.~Li, S.~Liu, J.~Yang, and M.-H. Yang, ``Generative face completion,'' in
  \emph{Proceedings of the IEEE conference on computer vision and pattern
  recognition}, 2017, pp. 3911--3919.

\bibitem{xiao2021face}
J.~Xiao, D.~Zhan, H.~Qi, and Z.~Jin, ``When face completion meets irregular
  holes: An attributes guided deep inpainting network,'' in \emph{Proceedings
  of the 29th ACM International Conference on Multimedia}, 2021, pp.
  3202--3210.

\bibitem{eye}
B.~Dolhansky and C.~C. Ferrer, ``Eye in-painting with exemplar generative
  adversarial networks,'' in \emph{Proceedings of the IEEE conference on
  computer vision and pattern recognition}, 2018, pp. 7902--7911.

\bibitem{li2020learning}
X.~Li, G.~Hu, J.~Zhu, W.~Zuo, M.~Wang, and L.~Zhang, ``Learning symmetry
  consistent deep cnns for face completion,'' \emph{IEEE Transactions on Image
  Processing}, vol.~29, pp. 7641--7655, 2020.

\bibitem{zhou2020learning}
T.~Zhou, C.~Ding, S.~Lin, X.~Wang, and D.~Tao, ``Learning oracle attention for
  high-fidelity face completion,'' in \emph{Proceedings of the IEEE/CVF
  Conference on Computer Vision and Pattern Recognition}, 2020, pp. 7680--7689.

\bibitem{sun2005image}
J.~Sun, L.~Yuan, J.~Jia, and H.-Y. Shum, ``Image completion with structure
  propagation,'' in \emph{ACM SIGGRAPH 2005 Papers}, 2005, pp. 861--868.

\bibitem{pavic2006interactive}
D.~Pavi{\'c}, V.~Sch{\"o}nefeld, and L.~Kobbelt, ``Interactive image completion
  with perspective correction,'' \emph{The Visual Computer}, vol.~22, no.~9,
  pp. 671--681, 2006.

\bibitem{hays2007scene}
J.~Hays and A.~A. Efros, ``Scene completion using millions of photographs,''
  \emph{ACM Transactions on Graphics (ToG)}, vol.~26, no.~3, pp. 4--es, 2007.

\bibitem{huang2013transformation}
J.-B. Huang, J.~Kopf, N.~Ahuja, and S.~B. Kang, ``Transformation guided image
  completion,'' in \emph{IEEE International Conference on Computational
  Photography (ICCP)}.\hskip 1em plus 0.5em minus 0.4em\relax IEEE, 2013, pp.
  1--9.

\bibitem{jo2019sc}
Y.~Jo and J.~Park, ``Sc-fegan: face editing generative adversarial network with
  user's sketch and color,'' in \emph{Proceedings of the IEEE/CVF International
  Conference on Computer Vision}, 2019, pp. 1745--1753.

\bibitem{zhang2020text}
L.~Zhang, Q.~Chen, B.~Hu, and S.~Jiang, ``Text-guided neural image
  inpainting,'' in \emph{Proceedings of the 28th ACM International Conference
  on Multimedia}, 2020, pp. 1302--1310.

\bibitem{Transfill}
Y.~Zhou, C.~Barnes, E.~Shechtman, and S.~Amirghodsi, ``Transfill:
  Reference-guided image inpainting by merging multiple color and spatial
  transformations,'' in \emph{Proceedings of the IEEE/CVF Conference on
  Computer Vision and Pattern Recognition}, 2021, pp. 2266--2276.

\bibitem{luo2022photo}
W.~Luo, S.~Yang, and W.~Zhang, ``Photo-realistic image synthesis from lines and
  appearance with modular modulation,'' \emph{Neurocomputing}, vol. 503, pp.
  81--91, 2022.

\bibitem{pix2pix}
P.~Isola, J.-Y. Zhu, T.~Zhou, and A.~A. Efros, ``Image-to-image translation
  with conditional adversarial networks,'' in \emph{Proceedings of the IEEE
  conference on computer vision and pattern recognition}, 2017, pp. 1125--1134.

\bibitem{HDpix2pix}
T.-C. Wang, M.-Y. Liu, J.-Y. Zhu, A.~Tao, J.~Kautz, and B.~Catanzaro,
  ``High-resolution image synthesis and semantic manipulation with conditional
  gans,'' in \emph{Proceedings of the IEEE conference on computer vision and
  pattern recognition}, 2018, pp. 8798--8807.

\bibitem{SPADE}
T.~Park, M.-Y. Liu, T.-C. Wang, and J.-Y. Zhu, ``Semantic image synthesis with
  spatially-adaptive normalization,'' in \emph{Proceedings of the IEEE
  Conference on Computer Vision and Pattern Recognition}, 2019, pp. 2337--2346.

\bibitem{SEAN}
P.~Zhu, R.~Abdal, Y.~Qin, and P.~Wonka, ``Sean: Image synthesis with semantic
  region-adaptive normalization,'' in \emph{Proceedings of the IEEE/CVF
  Conference on Computer Vision and Pattern Recognition}, 2020, pp. 5104--5113.

\bibitem{li2019faceshifter}
L.~Li, J.~Bao, H.~Yang, D.~Chen, and F.~Wen, ``Advancing high fidelity identity
  swapping for forgery detection,'' in \emph{Proceedings of the IEEE/CVF
  Conference on Computer Vision and Pattern Recognition (CVPR)}, June 2020.

\bibitem{jamaludin2019you}
A.~Jamaludin, J.~S. Chung, and A.~Zisserman, ``You said that?: Synthesising
  talking faces from audio,'' \emph{International Journal of Computer Vision},
  vol. 127, no.~11, pp. 1767--1779, 2019.

\bibitem{prajwal2020lip}
K.~Prajwal, R.~Mukhopadhyay, V.~P. Namboodiri, and C.~Jawahar, ``A lip sync
  expert is all you need for speech to lip generation in the wild,'' in
  \emph{Proceedings of the 28th ACM International Conference on Multimedia},
  2020, pp. 484--492.

\bibitem{duan2020look}
Q.~Duan and L.~Zhang, ``Look more into occlusion: Realistic face frontalization
  and recognition with boostgan,'' \emph{IEEE transactions on neural networks
  and learning systems}, vol.~32, no.~1, pp. 214--228, 2020.

\bibitem{liu2021blendgan}
M.~Liu, Q.~Li, Z.~Qin, G.~Zhang, P.~Wan, and W.~Zheng, ``Blendgan: Implicitly
  gan blending for arbitrary stylized face generation,'' \emph{Advances in
  Neural Information Processing Systems}, vol.~34, 2021.

\bibitem{unet}
O.~Ronneberger, P.~Fischer, and T.~Brox, ``U-net: Convolutional networks for
  biomedical image segmentation,'' in \emph{International Conference on Medical
  image computing and computer-assisted intervention}.\hskip 1em plus 0.5em
  minus 0.4em\relax Springer, 2015, pp. 234--241.

\bibitem{yan2018shift}
Z.~Yan, X.~Li, M.~Li, W.~Zuo, and S.~Shan, ``Shift-net: Image inpainting via
  deep feature rearrangement,'' in \emph{Proceedings of the European conference
  on computer vision (ECCV)}, 2018, pp. 1--17.

\bibitem{deng2019arcface}
J.~Deng, J.~Guo, N.~Xue, and S.~Zafeiriou, ``Arcface: Additive angular margin
  loss for deep face recognition,'' in \emph{Proceedings of the IEEE/CVF
  Conference on Computer Vision and Pattern Recognition}, 2019, pp. 4690--4699.

\bibitem{faceparsing}
zllrunning, ``face-parsing.pytorch,''
  \url{https://github.com/zllrunning/face-parsing.PyTorch}, 2019.

\bibitem{resnet}
K.~He, X.~Zhang, S.~Ren, and J.~Sun, ``Deep residual learning for image
  recognition,'' in \emph{Proceedings of the IEEE conference on computer vision
  and pattern recognition}, 2016, pp. 770--778.

\bibitem{johnson2016perceptual}
J.~Johnson, A.~Alahi, and L.~Fei-Fei, ``Perceptual losses for real-time style
  transfer and super-resolution,'' in \emph{European conference on computer
  vision}.\hskip 1em plus 0.5em minus 0.4em\relax Springer, 2016, pp. 694--711.

\bibitem{vgg}
K.~Simonyan and A.~Zisserman, ``Very deep convolutional networks for
  large-scale image recognition,'' \emph{arXiv preprint arXiv:1409.1556}, 2014.

\bibitem{FCN}
J.~Long, E.~Shelhamer, and T.~Darrell, ``Fully convolutional networks for
  semantic segmentation,'' in \emph{Proceedings of the IEEE conference on
  computer vision and pattern recognition}, 2015, pp. 3431--3440.

\bibitem{brock2018large}
A.~Brock, J.~Donahue, and K.~Simonyan, ``Large scale gan training for high
  fidelity natural image synthesis,'' in \emph{International Conference on
  Learning Representations}, 2018.

\bibitem{liu2019few}
M.-Y. Liu, X.~Huang, A.~Mallya, T.~Karras, T.~Aila, J.~Lehtinen, and J.~Kautz,
  ``Few-shot unsupervised image-to-image translation,'' in \emph{Proceedings of
  the IEEE/CVF International Conference on Computer Vision}, 2019, pp.
  10\,551--10\,560.

\bibitem{FID}
M.~Heusel, H.~Ramsauer, T.~Unterthiner, B.~Nessler, and S.~Hochreiter, ``Gans
  trained by a two time-scale update rule converge to a local nash
  equilibrium,'' in \emph{Advances in Neural Information Processing Systems},
  2017, pp. 6626--6637.

\bibitem{szegedy2016rethinking}
C.~Szegedy, V.~Vanhoucke, S.~Ioffe, J.~Shlens, and Z.~Wojna, ``Rethinking the
  inception architecture for computer vision,'' in \emph{Proceedings of the
  IEEE conference on computer vision and pattern recognition}, 2016, pp.
  2818--2826.

\bibitem{LPIPS}
R.~Zhang, P.~Isola, A.~A. Efros, E.~Shechtman, and O.~Wang, ``The unreasonable
  effectiveness of deep features as a perceptual metric,'' in \emph{Proceedings
  of the IEEE conference on computer vision and pattern recognition}, 2018, pp.
  586--595.

\bibitem{wang2018cosface}
H.~Wang, Y.~Wang, Z.~Zhou, X.~Ji, D.~Gong, J.~Zhou, Z.~Li, and W.~Liu,
  ``Cosface: Large margin cosine loss for deep face recognition,'' in
  \emph{Proceedings of the IEEE conference on computer vision and pattern
  recognition}, 2018, pp. 5265--5274.

\bibitem{adam}
D.~P. Kingma and J.~Ba, ``Adam: A method for stochastic optimization,'' in
  \emph{ICLR (Poster)}, 2015.

\bibitem{comod}
S.~Zhao, J.~Cui, Y.~Sheng, Y.~Dong, X.~Liang, I.~Eric, C.~Chang, and Y.~Xu,
  ``Large scale image completion via co-modulated generative adversarial
  networks,'' in \emph{International Conference on Learning Representations},
  2020.

\bibitem{faceswap}
deepfakes, ``faceswap,'' \url{https://github.com/deepfakes/faceswap}, 2018.

\bibitem{DeepFaceLab}
iperov, ``Deepfacelab,''
  \url{https://github.com/zllrunning/face-parsing.PyTorch}, 2018.

\bibitem{nirkin2019fsgan}
Y.~Nirkin, Y.~Keller, and T.~Hassner, ``{FSGAN}: Subject agnostic face swapping
  and reenactment,'' in \emph{Proceedings of the IEEE International Conference
  on Computer Vision}, 2019, pp. 7184--7193.

\bibitem{liu2015faceattributes}
Z.~Liu, P.~Luo, X.~Wang, and X.~Tang, ``Deep learning face attributes in the
  wild,'' in \emph{Proceedings of International Conference on Computer Vision
  (ICCV)}, December 2015.

\bibitem{InsightFace-v2}
foamliu, ``Insightface-v2,'' \url{https://github.com/foamliu/InsightFace-v2},
  2019.

\end{thebibliography}

\vfill


\end{document}